\documentclass[10pt,conference]{IEEEtran}
\usepackage{cite}
\usepackage{amsmath,amssymb,amsfonts}
\usepackage{algorithmic}
\usepackage{graphicx}
\usepackage{textcomp}
\usepackage[table]{xcolor}
\usepackage[hyphens]{url}
\usepackage{fancyhdr}
\usepackage{hyperref}
\usepackage{comment}
\usepackage{multirow}
\usepackage{booktabs}
\usepackage{arydshln}
\usepackage{xspace}
\newif\ifshowcomment
\showcommentfalse

\ifshowcomment
    \newcommand{\chaojie}[1]{{\color{magenta}[CZ: #1]}}
    \newcommand{\jovan}[1]{{\color{olive}[JS: #1]}}
    \newcommand{\todo}[1]{{\color{red}[TODO: #1]}}
\else
    \newcommand{\chaojie}[1]{\ignorespaces}
    \newcommand{\jovan}[1]{\ignorespaces}
    \newcommand{\todo}[1]{\ignorespaces}
\fi
\newcommand{\TODO}[1]{}
\newcommand{\system}{Agora}
\newcommand{\company}{\emph{Microsoft Azure}}

\title{Architectural Implications of Agentic AI Workflows}

\newcommand\paperauthors{Jirong Yang$^{\dagger}$, Peizhe Liu$^{\dagger}$, Chaojie Zhang$^{*}$, Jovan Stojkovic$^{\dagger}$}
\newcommand\paperaffiliation{$^{\dagger}$The University of Texas at Austin \qquad $^{*}$Microsoft Azure}

\author{
  \IEEEauthorblockN{\paperauthors{}}
  \IEEEauthorblockA{\paperaffiliation{}}
}

\begin{document}
\maketitle

\thispagestyle{plain}
\pagestyle{plain}

\newcommand{\myparagraph}[1]{\vspace{1pt}\noindent\textbf{#1.\xspace}}


\begin{abstract}
Agentic AI, where LLMs autonomously plan, invoke tools, and
coordinate across agents to solve complex tasks, is an emerging
class of datacenter workloads. However, its architectural implications remain
largely unexplored. To close this gap, we organize agentic workflows through a taxonomy and present the first architectural characterization
of agentic AI using both a production study across the fleet of our hyperscaler and a controlled study of representative open-source frameworks.

Our study shows that agentic execution is fundamentally fragmented and heterogeneous.
Each request expands into a workflow of LLM inferences, tool invocations, and
orchestration decisions that repeatedly crosses the CPU--GPU boundary. Our taxonomy
explains how this fragmentation turns into resource demand. As orchestration and
tools run on the host, the CPU sits on the critical path. The execution structure sets
the load over time, which stays low with sudden bursts when a stage
releases its tools. The model composition sets how evenly the workflow uses the GPUs,
often saturating some while leaving others idle. Diversity in tasks and tools widens
this range even further. 
These characteristics expose three architectural mismatches of conventional uniform servers. Fragmented execution strands CPU and GPU capacity despite bursty demand. Different software roles make homogeneous CPU provisioning inefficient. Finally, multiplexing many agents onto shared cores degrades microarchitectural locality and increases coordination overhead.

Guided by these findings, we derive implications for agentic servers and examine
them through \system{}, our prototype for commodity servers. 
\system{} dynamically harvests idle CPU cores for
co-located throughput work, while protecting agentic tail latency against tool spikes. It also oversubscribes GPU memory by placing more agents on each GPU, prefetching the
next agent's state to hide swap latency. To match the machine to the heterogeneous
roles, \system{} pools cores by role and applies affinity-aware scheduling to restore
locality. Finally, \system{} automatically tunes all of these mechanisms to the 
running workload.
These techniques substantially improve
CPU and GPU utilization and per-server throughput while preserving agent tail latency. 
Our insights also identify key directions for
future server architectures for agentic AI. 
\end{abstract}

\vspace{0mm}
\section{Introduction}
\label{sec:intro}

The way large language models (LLMs) are deployed is changing. Traditionally, LLM serving systems have optimized each model invocation as an independent request: the system consumes a prompt and produces a response~\cite{orca,vllm,flashattention,spotserve,pets,splitwise,alizadeh2024llm,alisa}.
However, a growing class of applications now embeds one or more models in an iterative control loop. At each step, the application uses model inference to interpret the current state, select the next action, such as invoking a tool or delegating a subtask, observe the result, and continue until the task is complete.
This \emph{agentic} mode of execution powers applications in
software engineering~\cite{copilot}, deep research~\cite{cheng2025barbariansgateaiupending},
scientific discovery~\cite{alphaevolve}, and enterprise automation~\cite{ibmagenticenterprise}.
Projections place it among the fastest-growing consumers of
datacenter capacity in the coming years~\cite{agenticmarket}.

Despite this popularity, the implications of agentic AI for server architecture remain underexplored. Researchers have been optimizing servers for traditional, CPU-centric services such as web
serving, key-value stores, and data analytics~\cite{micromanycore,hardharvest,mosaic,duplexity,phaseweave,rpcvalet,nanopu,accelerometer,accelflow,protoAcc,cdpu,warehouseScaleComp,deathstarbench,dcperf,archimplrec}, and, more recently, the
inference of a single monolithic LLM, where dense tensor computation on an
accelerator dominates while the host merely feeds it~\cite{splitwise,alisa,alizadeh2024llm,llmopt1,llmopt2,llmopt4,llmopt5}.
Agentic execution differs from both. A task no longer maps to a single 
tensor computation or a request-response service. Instead, it unfolds
as a data-dependent graph of many model calls, tool invocations,
and control decisions that orchestration software assembles on the host.
Thus, execution spans the accelerator and multiple host-side components (CPU, memory system, network, and runtime) and grows complex as applications add agents and deepen their coordination.

Although a nascent body of systems work has begun to build agentic
frameworks~\cite{yu2026pythia, kim2025cost, kang2026thunderagent, bian2026tokendance, murakkab}, we lack fundamental architectural understanding of how
these workloads exercise real hardware 
and whether
servers designed for the two older targets are adequate. 
To reason systematically about such highly diverse workloads, we first develop a taxonomy along platform-relevant dimensions of agentic AI: how agents are orchestrated, how their workflows are
structured, and how models are composed. 
Guided by the taxonomy, we perform the first architectural characterization of agentic AI through both a production study across the fleet of our hyperscaler, \company{}, and a controlled study of diverse open-source frameworks, spanning diverse use cases. 

Our main insight is that agentic AI creates highly fragmented
execution, while the dimensions in our taxonomy determine how this
fragmentation manifests as heterogeneous resource demand. Since host-side orchestration and tools separate successive model calls, the CPU enters the critical path and each task repeatedly crosses the CPU--GPU boundary.
The orchestration mechanism governs how control passes among agents.
Execution structure shapes the 
pattern of host load: long periods of low utilization can be punctuated
by bursts toward saturation when workflow stages release groups of tool calls.
Model composition determines how evenly a workflow utilizes the
accelerator pool, potentially saturating some GPUs while leaving others idle.
Even within the same taxonomy class, workloads vary substantially in offered
load and the tasks they solve, further widening the range of
resource needs. 

Consequently, conventional, uniformly provisioned servers are ill-suited
to agentic AI in three ways. First, fragmentation strands resources: CPUs and
GPUs remain underutilized on average yet experience short bursts of
near-saturation, leaving substantial capacity idle between bursts. Second, the
three host roles (schedulers, orchestrators, and runners) exhibit sharply
different resource profiles, so a single homogeneous pool of cores cannot serve
all of them efficiently. Third, multiplexing many agents on shared cores
degrades microarchitectural locality, as they evict one another's cache lines
and branch-predictor state, increasing pipeline stalls. As the nature and
severity of these inefficiencies vary with the workflow, load, and task, no
single static configuration suits them all; the server must instead adapt
to each workload.

Guided by these findings, we derive three design principles for servers running agentic
AI.  
We examine each through a case study implemented in \emph{\system{}}, our
prototype for commodity servers. First, on the host side, servers should reclaim the temporal slack from fragmented execution. 
\system{} harvests idle CPU cores for co-located throughput work while protecting
agents from sudden tool bursts. Second, on the accelerator side, servers should reclaim the idle capacity rather than dedicating exclusive GPU access to fixed agents. \system{}
harvests this capacity by consolidating agents onto fewer GPUs, exploiting agents that
share model state or are never active at the same time so that busy and idle devices
are balanced. Third, host cores should be partitioned by software roles and scheduled to
preserve task locality, rather than managed as a single homogeneous pool. \system{}
isolates and right-sizes the control plane and the bursty runner pool, and pins tasks
within the runner pool to preserve cache and branch-predictor locality. 

We implement the three case studies on open-source agentic frameworks and
evaluate them on real hardware against static, workload-agnostic configurations.
CPU harvesting recovers $95\%$ of a co-located workload's
standalone throughput and increases host CPU utilization by $30\%$ while limiting agent slowdown to under $3\%$. 
GPU harvesting
frees a third of the GPUs 
while raising generation throughput by $82\%$ and cutting tail
latency by $2.5\times$. Role-aware pooling reduces
the tools' CPU demand by up to $46\%$ 
and worst-case tool latency by $13\%$,
while retaining $99\%$ of serving throughput.

Beyond today's hardware, our insights also suggest optimizations for future hardware. Dedicated support could offload the frequent
scheduling and context-management operations off the host cores. 
CPU cores could further match hardware
capabilities to different host roles, combining energy-efficient cores for
lightweight coordination with high-performance cores for compute-intensive tool
execution.
 
In summary, this paper makes the
following contributions:
\begin{itemize}
  \item The first production architectural characterization of agentic AI, measured across the fleet of a hyperscaler.
  
  \item A taxonomy that organizes the heterogeneous space of agentic AI workloads along platform-relevant axes, with a characterization of real use cases across its classes.

  \item Three server-design implications, explored through case studies in \system{}:
  harvesting idle CPU cores, harvesting idle GPU capacity by consolidating agents, and
  managing host cores by role and task locality.

\end{itemize}

\section{Background and Motivation}
\label{sec:background}

\begin{figure}[t]
  \centering
  \includegraphics[width=\columnwidth]{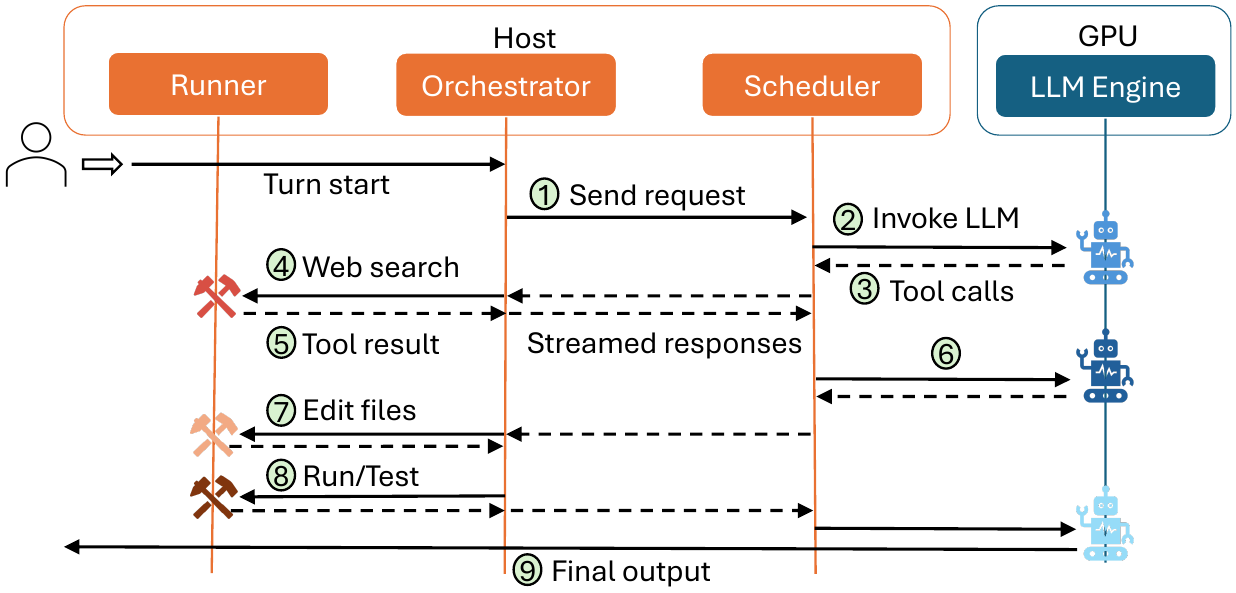}
  \vspace{-7mm}
  \caption{An example of agentic AI workflow execution.  
  }
  \label{fig:agentic-overview}
  \vspace{-4mm}
\end{figure}

\definecolor{taxhead}{HTML}{34495E}   
\definecolor{taxopta}{HTML}{D6E4F0}   
\definecolor{taxoptb}{HTML}{EEF4FA}   
\definecolor{taxrule}{HTML}{7F8FA6}   

\begin{table*}[t]
\centering
\caption{Taxonomy of agentic workflows with examples that fall within different categories and their implications for platform design.  
}
\vspace{-2mm}
\label{tab:taxonomy}
\setlength{\tabcolsep}{6pt}
\renewcommand{\arraystretch}{1.3}
\arrayrulecolor{taxrule}
\begin{tabular}{p{0.14\textwidth} p{0.10\textwidth} p{0.33\textwidth} p{0.33\textwidth}}
\rowcolor{taxhead}
\textcolor{white}{\textbf{Dimension}} & \textcolor{white}{\textbf{Option}} &
\textcolor{white}{\textbf{Examples in the current corpus}} & \textcolor{white}{\textbf{Implication for platform design}} \\
\rowcolor{taxopta}
\multirow{2}{0.15\textwidth}{\textbf{Orchestration}} &
Host-orchestrated & MetaGPT~\cite{metagpt}; MAGE~\cite{zhao2024mage}; Paper2Code~\cite{paper2code} &
The host exposes the control flow to the runtime, enabling scheduling and prefetching. \\
\cdashline{2-4}
\rowcolor{taxoptb}
& LLM-orchestrated & OWL~\cite{hu2025owloptimizedworkforcelearning}; ToolOrchestra~\cite{toolorchestra}; AOrchestra~\cite{aorchestra} &
The model selects or creates agents at runtime, adding inference-driven control decisions. \\
\midrule
\rowcolor{taxopta}
\multirow{2}{0.15\textwidth}{\textbf{Execution structure}} &
Sequential & AlphaEvolve~\cite{alphaevolve}; Trae Agent~\cite{traeresearchteam2025traeagentllmbasedagent}; AccelOpt~\cite{accelopt} &
Dependent steps form a critical path; a stalled step delays the whole workflow. \\
\cdashline{2-4}
\rowcolor{taxoptb}
& Parallel & ReConcile~\cite{chen2024reconcile}; CORAL~\cite{qu2026coralautonomousmultiagentevolution}; DeLM~\cite{mao2026delm} &
Independent agents create concurrent inference and tool bursts, increasing batching and load. \\
\midrule
\rowcolor{taxopta}
\multirow{2}{0.15\textwidth}{\textbf{Model composition}} &
Homogeneous & Most frameworks use the same base model by default &
A common base model enables shared residency and larger cross-agent batches. \\
\cdashline{2-4}
\rowcolor{taxoptb}
& Heterogeneous & ToolOrchestra~\cite{toolorchestra}; AOrchestra~\cite{aorchestra}; AccelOpt~\cite{accelopt} &
Multiple model types must remain available within the workflow, complicating load balancing. \\
\end{tabular}
\arrayrulecolor{black}
\vspace{-5mm}
\end{table*}

\noindent \textbf{Agentic AI Workflows.}
\label{subsec:workflows}
Traditional LLM serving treats model invocation as the primary
unit of execution. A serving engine prompts the model with context, decodes output tokens autoregressively, and returns a
response~\cite{orca,vaswani2017attention}.
The host prepares, schedules, and dispatches
requests, while the accelerator, e.g., GPU, executes model computation \cite{vllm,tensorRTLLM,zheng2024sglangefficientexecutionstructured}.

An \emph{agent} places an LLM inside a control loop. Given a goal, it invokes the model to interpret the current state,
plan or reflect, and select a next step. The step may invoke an external \emph{tool}, such as web search, code execution, or retrieval~\cite{websearchopenAI,dbsagenticAWS}, whose result is included in next model calls~\cite{yao2023reactsynergizingreasoningacting}.
Tools run off the accelerator, on the host or a remote service~\cite{yao2023reactsynergizingreasoningacting,kim2025cost}. 

As each action depends on the previous one, a user request expands into a
\emph{workflow}: a data-dependent graph whose nodes are model inferences, tool
invocations, and control decisions, and whose edges are their
dependencies~\cite{langgraph,autogen}. The number, duration, and dependencies of these
steps are determined dynamically at runtime, so execution alternates between accelerator-bound
generation and host-bound work that tracks dependencies, marshals state, and invokes
tools. Figure~\ref{fig:agentic-overview} shows this fragmented, host-orchestrated execution. 

\emph{Multi-agent} applications distribute this workflow across cooperating
agents. 
A coordinator may
assign subtasks to specialized agents, e.g.,  researchers, programmers, reviewers
\cite{ibmwhatismultiagent,crewAI}. Agents may exchange results and claim
work through messages or shared state. Each maintains its own logical control loop, while multiple
agents may share the same model backend.
Independent subtasks can execute concurrently, while dependencies and iterative feedback
serialize others.

\vspace{2pt}
\noindent \textbf{Agentic Serving Systems.}
\label{subsec:agentic-serving}
Researchers have been extensively optimizing LLM inference serving. Continuous batching~\cite{orca},
KV-cache management~\cite{vllm}, optimized attention kernels~\cite{flashattention,zheng2024sglangefficientexecutionstructured}, disaggregated prefill and
decode~\cite{splitwise}, all raise the throughput of the inference
engine. These techniques optimize the inference-serving plane. Agentic workflows open a richer optimization space defined by dependencies among model invocations and tool calls. A growing body of
systems work exploits this structure. Speculative agent execution runs
likely-next actions ahead of time and discards
mispredictions~\cite{ye2026speculativeactionslosslessframework}. Programmable,
software-defined serving lets applications express and co-optimize the workflow
itself~\cite{agarwal2026softwaredefinedagenticserving}. Agent-aware schedulers
efficiently order workflow steps~\cite{agentix,kang2026thunderagent, yu2026pythia}.

\vspace{2pt}
\noindent \textbf{Agentic CPU.}
\label{subsec:agentic-cpu} 
The 
aforementioned software optimizations 
still run on unchanged hardware.
Agentic execution changes the demands placed on this hardware, particularly on the host CPU. In LLM inference the CPU fills a single role:
a \emph{scheduler} that dispatches inferences
to the accelerator. 
Agentic execution keeps this role and adds two
additional logical host roles: an
\emph{orchestrator} that routes messages and state among agents and enforces
workflow dependencies, and a \emph{runner} that executes agent logic and invokes
tools~\cite{langgraph,autogen}. 
Thus, the CPU becomes far more than a front end for the accelerator. 
Figure~\ref{fig:agentic-overview} shows how these
three roles interleave CPU and GPU work across a workflow. 
This expansion has led several vendors to advertise ``agentic'' CPUs
and platforms~\cite{armagicpu,amdagenticcpu,agenticmorecpus}. Yet the architectural requirements of such platforms remain unclear.
The research community has also begun to examine the CPU in agentic execution, but
primarily at a coarse granularity, reporting aggregate host utilization and end-to-end
latency~\cite{raj2025cpucentricperspectiveagenticai,yuan2026agentic}. It remains unknown where the host
bottlenecks lie, how they shift across the three roles, what
microarchitectural challenges agentic workloads pose, and what server mechanisms these behaviors call for. 
\section{A Taxonomy of Agentic Workflows}
\label{sec:taxonomy}

Agentic workflows differ widely across deployments, making it difficult to reason about them as a single class. 
We take an \emph{agent}, a multi-turn model--tool loop, as the primitive of analysis, 
and present a taxonomy of multi-agent workflows. Interactions among such primitives expose platform-level choices and trade-offs.
We organize the space along three orthogonal, platform-relevant dimensions, each
placing a distinct demand on the server: who decides what runs next
(\emph{orchestration}), how the steps are arranged in time (\emph{execution
structure}), and how many model families are used in the workflow
(\emph{model composition}). Table~\ref{tab:taxonomy} summarizes the taxonomy. 
These dimensions are not immutable labels, instead, a workflow may combine multiple options or move between them across stages~\cite{gottweis2026accelerating, prabhakar2025enterprise,ping2025verimoa,liao2025kernelevolve}.
Thus, the taxonomy captures the dominant behavior that determines server needs. 

\vspace{2pt}\noindent\textbf{Orchestration.}  
The first dimension is who decides what runs next.
In a \emph{host-orchestrated} workflow, such task is done by the host-side program logic. 
A programmatic graph
or state machine specifies the agents and their dependencies,
and the host advances execution according to this logic, as in
MetaGPT~\cite{metagpt} and later multi-agent systems~\cite{zhao2024mage,paper2code}.
In an \emph{LLM-orchestrated} workflow, a model makes routing
decisions at runtime and forwards the decision to the host, so the control flow itself includes model inference, as in 
OWL~\cite{hu2025owloptimizedworkforcelearning},  
AOrchestra~\cite{aorchestra}, and 
ToolOrchestra~\cite{toolorchestra}. 
As host-side control logic is visible to the runtime, it exposes
workflow dependencies and can enable earlier scheduling and prefetching~\cite{yu2026pythia}.
LLM orchestration instead places a model inference in the control path:  
the next step depends on generated output, increasing
inference demand and reducing predictability. 

\vspace{2pt}

\noindent\textbf{Execution structure.}
The second dimension is how the nodes of the workflow graph are arranged over time.
In a \emph{sequential} workflow, each agent depends on
the result of the preceding one: at most one node is active at a time, 
forming a chain 
\cite{alphaevolve,traeresearchteam2025traeagentllmbasedagent,accelopt}. A
\emph{parallel} workflow instead fans out independent subtasks that run concurrently
and later synchronize, as in multi-agent debate~\cite{chen2024reconcile}, CORAL
\cite{qu2026coralautonomousmultiagentevolution},
and DeLM 
\cite{mao2026delm}.
This structure shapes the temporal pattern of demand on the
host and accelerators. A sequential workflow exposes
little independent work, so a delayed
inference, tool call, or handoff directly extends its critical path.
A parallel workflow exposes multiple ready agents at once, creating
opportunities for batching and overlap but also concentrating
inference and tool activity into bursts that increase pressure on the system's resources.

\vspace{2pt}
\noindent\textbf{Model composition.}
The third dimension is whether the agents run the same model or a mix. 
In a \emph{homogeneous} workflow, all agents draw from the same base
model, varying only the prompt, tools, or a lightweight adapter, as most open-source frameworks do by
default. A \emph{heterogeneous} workflow combines models of different kinds or
sizes~\cite{toolorchestra,aorchestra,accelopt}, e.g., a small model that plans
or routes and a large one that writes code~\cite{hybridLLM,agent3,agent1}. 
This distinction directly affects model residency and serving efficiency.
A homogeneous workflow creates opportunities to share one resident model across agents and batch their requests, with only the per-agent context stored separately, simplifying provisioning and placement. A heterogeneous workflow keeps several models resident, raising GPU-memory pressure, fragmenting batching, and  complicating load balancing.

\section{Characterizing Agentic AI Workloads}
\label{sec:characterization}
\label{sec:prod-char}
\label{sec:methodology} 

To understand the architectural implications of agentic AI, we characterize it through
two complementary lenses.
First, we conduct a
\emph{fleet study} of production agentic services running in our
hyperscaler, \company{}, capturing their real workload, tool mix, and resource usage at scale.  
Second, we perform a \emph{controlled study} using
four representative open-source frameworks covering representative points in our taxonomy, 
allowing us to isolate root causes and reproduce them. 
 
For the fleet study, we collect a 24-hour trace of production agentic requests and
attribute time, tool activity, and CPU metrics to software roles 
(schedulers, orchestrators, and runners). 
For the controlled
study, we evaluate \emph{SWE-Agent}~\cite{sweagent} (a single-agent, sequential, tool-intensive coding workflow), 
\emph{Trae}~\cite{traeresearchteam2025traeagentllmbasedagent} (a sequential coding workflow with long tool chains and three model endpoints), \emph{CORAL}~\cite{qu2026coralautonomousmultiagentevolution} (a four-agent, parallel, homogeneous-model 
evolutionary workflow), and \emph{Owl}~\cite{hu2025owloptimizedworkforcelearning} (a heterogeneous workforce with a
coordinator, six text roles, and vision and speech models). We run them on a
server with a 96-core AMD EPYC 7V12 CPU and eight NVIDIA
A100 GPUs. We deploy one vLLM instance per agent role while varying concurrency from 1 to 32 tasks.  

Our studies lead to two findings. First, agentic AI creates highly
fragmented and heterogeneous execution across the CPU and GPU
(Section~\ref{subsec:cc-fragmented}). Second, this execution pattern
stresses the server architecture in ways that differ from both
traditional cloud services and monolithic LLM inference, manifesting
as resource imbalance, poor microarchitectural locality, and increasing
coordination overhead (Section~\ref{sec:uarch}).

\subsection{Agentic AI Creates Fragmented and Heterogeneous Work}
\label{subsec:cc-fragmented}

An agentic request unfolds as a graph of
interleaved model inferences, tool invocations, and orchestration decisions distributed across CPU and GPU.
This fragmented execution is diversified by workflow designs, application
tasks, and tool behaviors, resulting in highly heterogeneous resource needs.

\begin{figure}[h]
\vspace{0mm}
  \centering
  \includegraphics[width=\columnwidth]{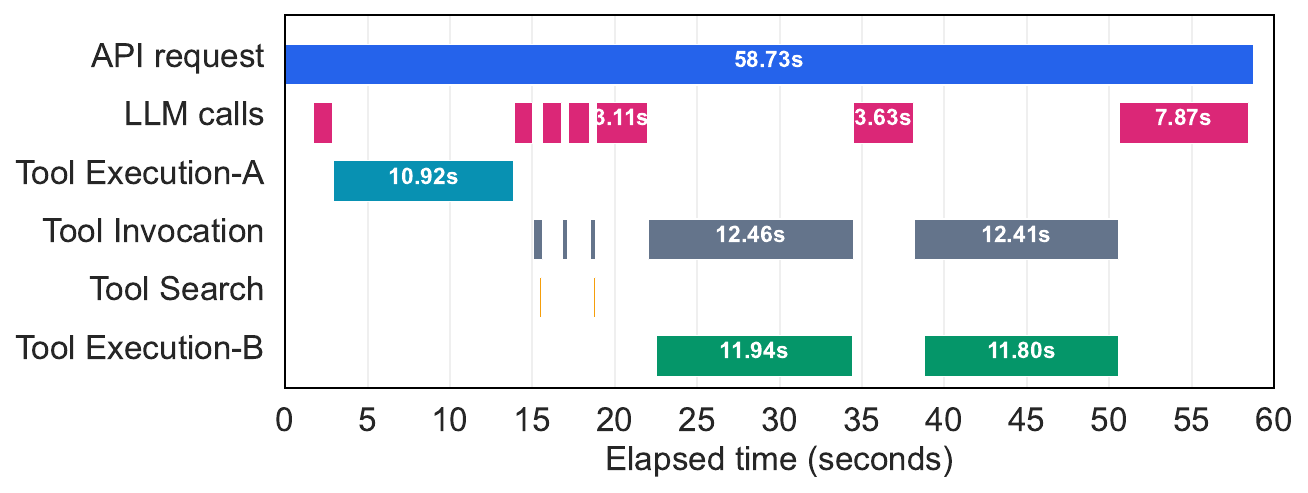}
  \vspace{-7mm}
  \caption{Lifecycle of an example production request. Tool phases
  interleave with LLM calls, and the request repeatedly crosses the CPU-GPU
  boundary.}
  \label{fig:cc-lifecycle}
  \vspace{-5mm}
\end{figure}

\myparagraph{Host orchestration fragments execution}
Figure~\ref{fig:cc-lifecycle} shows the lifecycle of
a representative production request.
The request spans nearly a minute, alternating between multiple LLM calls, agent hand-offs, 3 tool-discovery operations, and 3 tool executions across 2 different tools. 
After each LLM call, host-side orchestration interprets the
output and advances the workflow, while every tool result must be incorporated into workflow state before the agent resumes. Hence, the
request repeatedly crosses the CPU-GPU boundary, and the host-side tool phases
account for a substantial fraction of the end-to-end latency. The CPU therefore becomes part of the 
critical path for agentic requests instead of merely feeding the accelerator.

\begin{figure}[h]
\vspace{-2mm}
  \centering
  \includegraphics[width=\columnwidth]{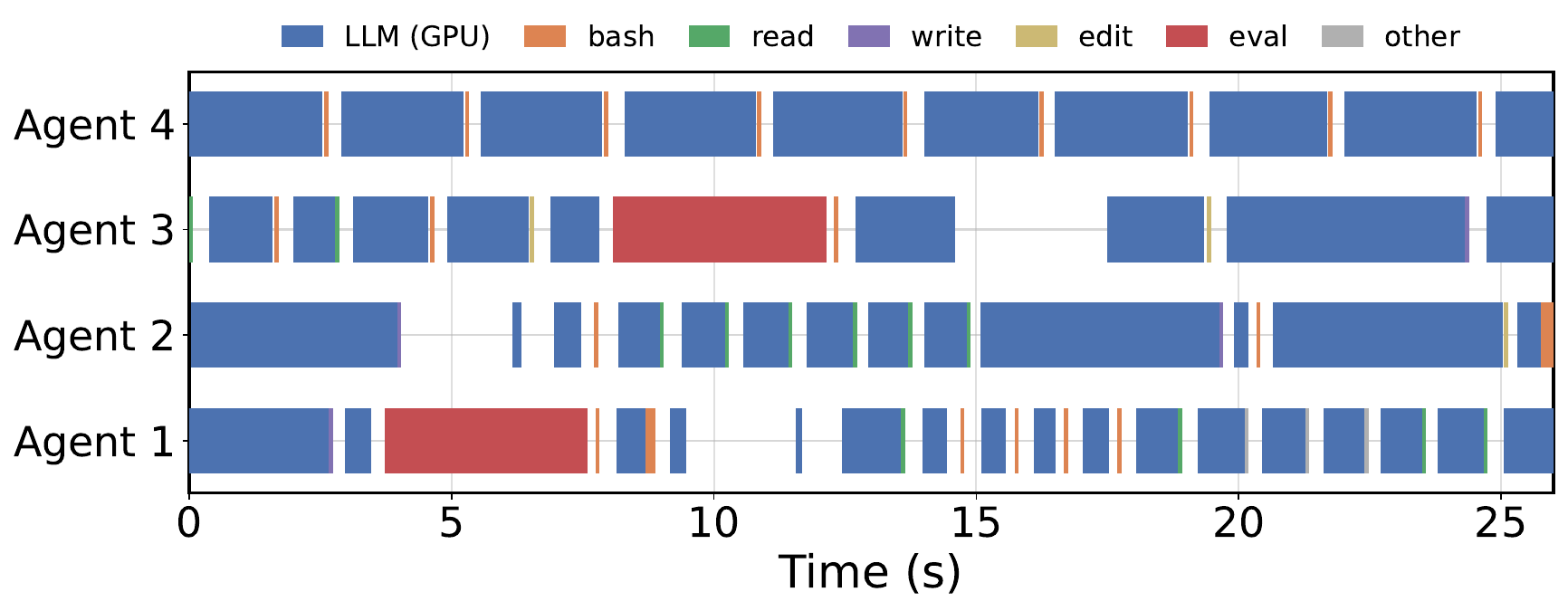}
  \vspace{-7mm}
  \caption{A slice of a CORAL run. Agents alternate LLM inferences (GPU)
  with tool calls (CPU), colored by tool type; dependencies leave idle gaps.}
  \label{fig:char-timeline}
  \vspace{-2mm}
\end{figure}

Figure~\ref{fig:char-timeline} shows the same behavior in our controlled study using CORAL.
A single run expands into $580$ LLM calls interleaved with $552$ heterogeneous tool invocations, ranging from millisecond file system operations to multi-second compilations and evaluations. 
Each agent alternates continuously between
GPU inference and CPU tool work, so execution ping-pongs between the two processors
hundreds of times. 
Control and data dependencies create visible gaps in an
agent's GPU activity while it waits for an upstream result. 
The fine-grained host-orchestrated execution fragments every request into heterogeneous CPU and GPU phases. 

\begin{figure}[h]
\vspace{-3mm}
  \centering
  \includegraphics[width=\columnwidth]{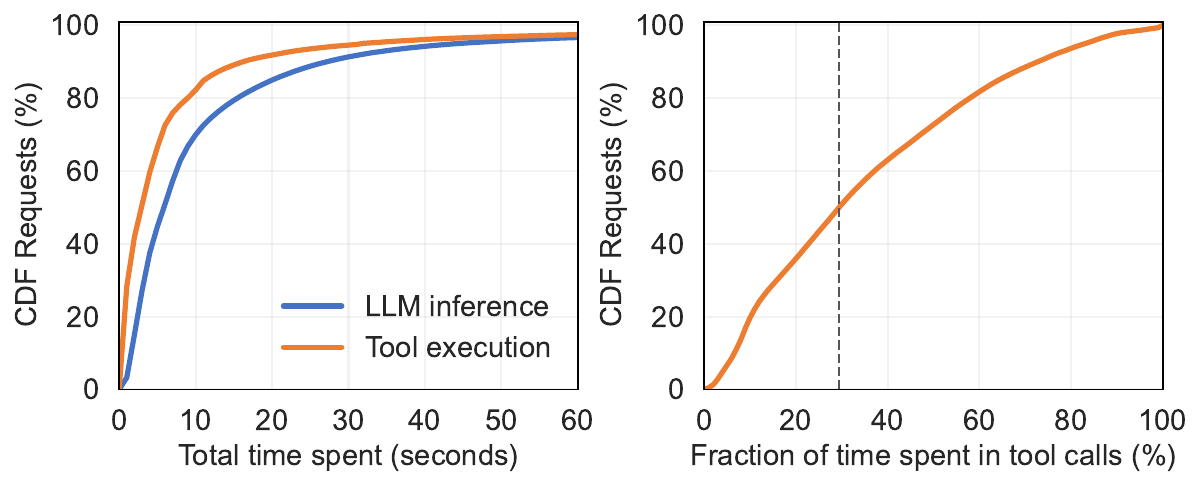}
  \vspace{-7mm}
  \caption{Fleet distribution of time spent in LLM inference versus tool execution,
  and the fraction of request time spent in tool calls.}
  \label{fig:cc-time}
  \vspace{-2mm}
\end{figure}

This pattern extends across the fleet. Figure~\ref{fig:cc-time} summarizes the
distribution of time agentic workflow requests spend in LLM inference
and tool execution, and the fraction of request time spent in tools. Tool execution time has a heavy tail distribution, accounting for a large and highly variable share of
request time, comparable to or exceeding the time spent in inference for more than 27\% of requests. 
Thus, host-side tool work is a first-class contributor to
end-to-end latency across the fleet, and the wide spread of the distribution means that no single
CPU-to-GPU time ratio characterizes the workload.

\myparagraph{Execution structure creates bursty load}
Execution structure shapes host load at multiple granularities.
At the agent level, sequential workflows expose one ready agent at
a time, while parallel workflows expose multiple agents
concurrently. Within an agent, a stage may further fan out into
multiple tool invocations. Both agent-level parallelism and
within-agent fan-out can concentrate host activity into short bursts.

\begin{figure}[h]
\vspace{-4mm}
  \centering
  \includegraphics[width=\columnwidth]{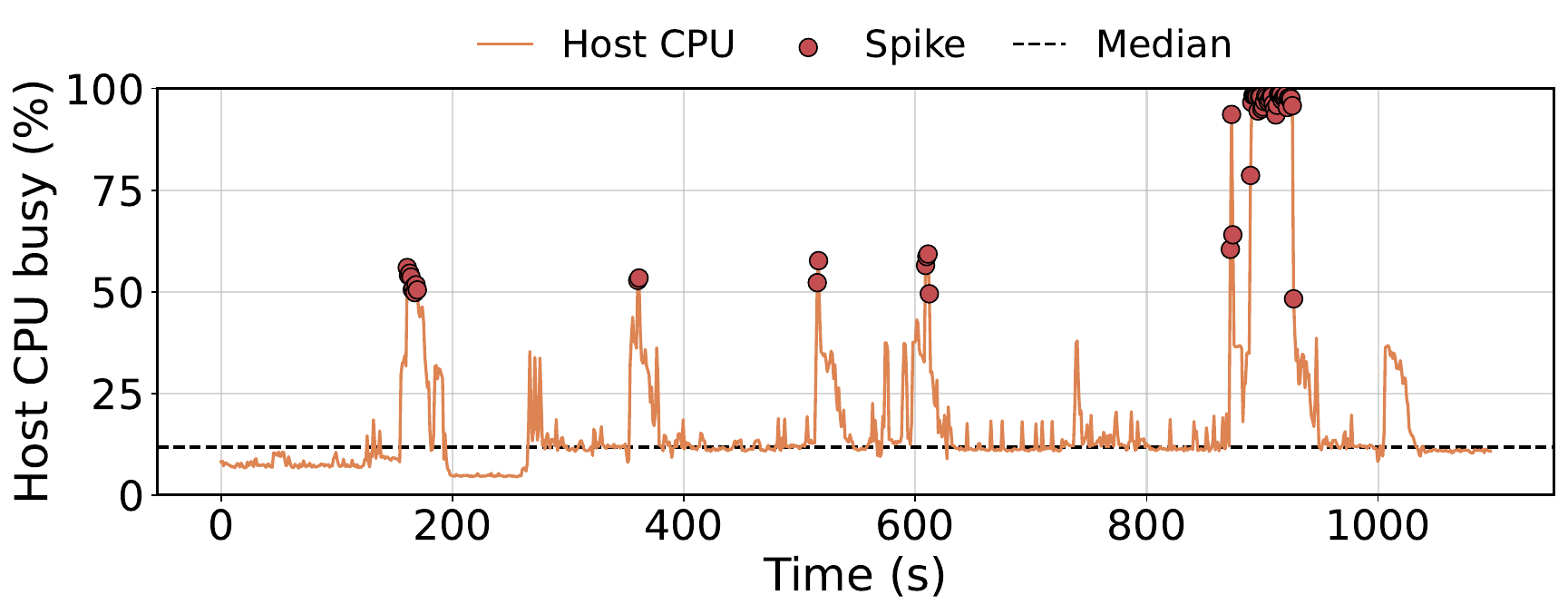}
  \vspace{-7mm}
  \caption{Host CPU over time for Trae. Stage-based execution keeps the CPU near
  its median, then drives sharp bursts toward saturation.}
  \label{fig:char-trae-cpu}
  \vspace{-3mm}
\end{figure}

Figure~\ref{fig:char-trae-cpu} shows the latter behavior through
host CPU utilization over an entire Trae run. Although Trae is
sequential at the agent level, some reasoning stages launch multiple
builds and test runs in parallel. Host CPU utilization stays near its
11\% median during the sequential portions, but rises rapidly to
nearly 100\% at these stage boundaries. This  contrast between
average and peak demand makes static provisioning inefficient. 

\begin{figure}[h]
\vspace{-3mm}
  \centering
  \includegraphics[width=\columnwidth]{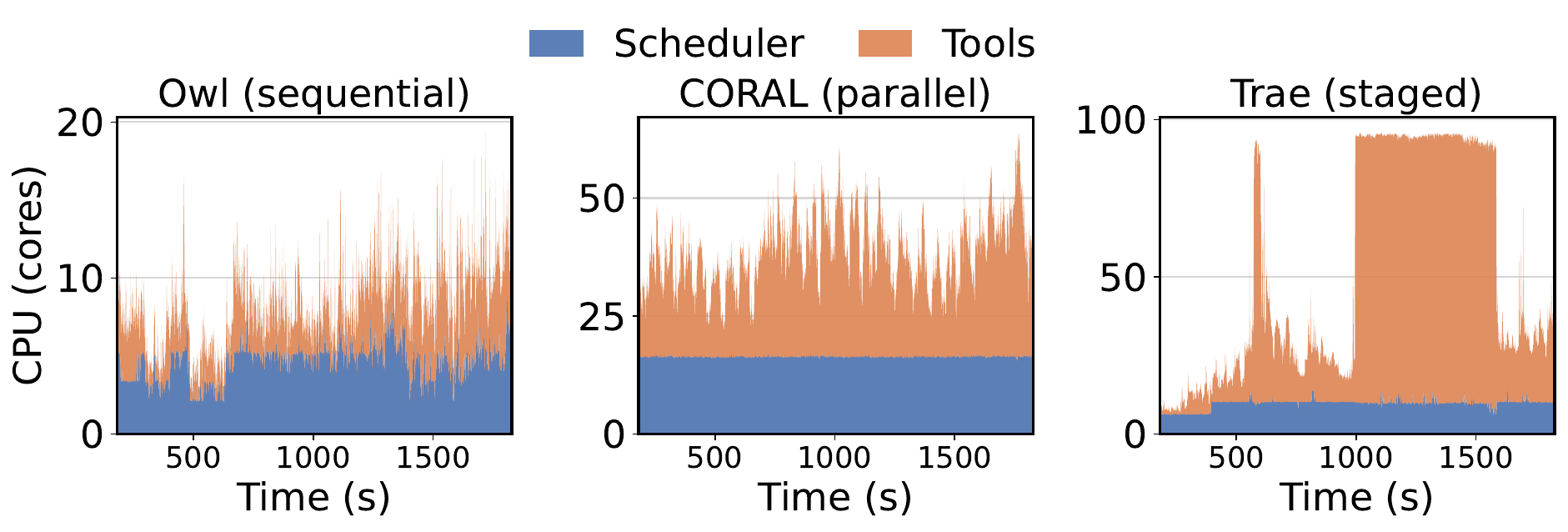}
  \vspace{-7mm}
  \caption{Host CPU over time split into scheduler cores (the per-agent serving
  engines) and tool cores for three different frameworks.}
  \label{fig:char-execstruct}
  \vspace{-3mm}
\end{figure}

Execution structure also shapes how host demand is divided between
scheduler and tool work. Under high concurrency, Figure~\ref{fig:char-execstruct} decomposes
host CPU demand over time into scheduler and tool components.
Scheduler demand remains relatively steady while an agent is active,
whereas tool demand fluctuates as tools start and finish.
The observed patterns align with each workflow's execution structure.
Owl activates only a subset of its roles at a time, so aggregate
scheduler demand remains low but fluctuates as different roles become
active and idle. CORAL runs its four agents in parallel, keeping
aggregate scheduler demand steadily high. Trae advances in stages,
causing tool demand to alternate between quiet intervals and extended,
near-saturating bursts.

\myparagraph{Model composition complicates imbalances across devices}
Model composition shapes how execution is distributed across
accelerators. In a homogeneous workflow, multiple roles may
share one serving instance or use role-dedicated instances of the
same model, whereas a heterogeneous workflow distributes execution
across models of different types or modalities.

\begin{figure}[h]
\vspace{0mm}
  \centering
  \includegraphics[width=\columnwidth]{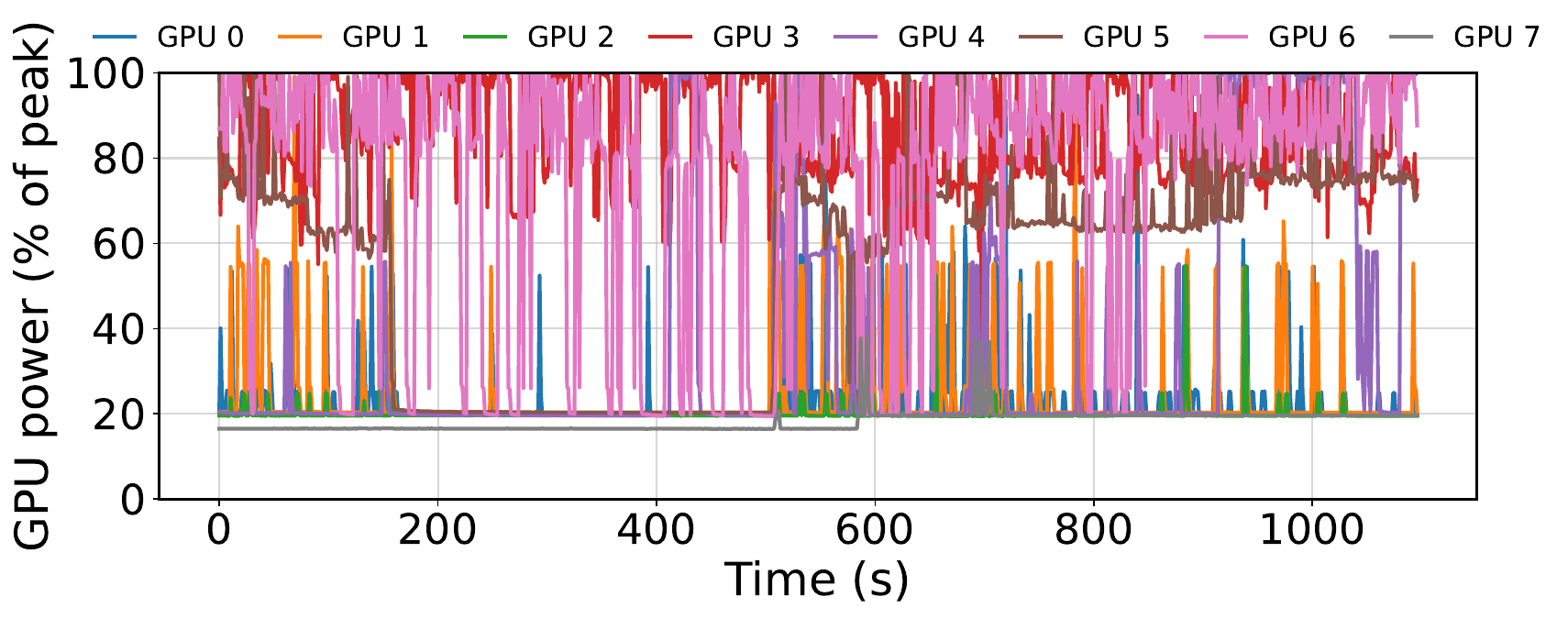}
  \vspace{-8mm}
  \caption{Per-GPU power consumption (\% of peak) over time for Owl.}
  \label{fig:char-owl-gpu}
  \vspace{-5mm}
\end{figure}

Figure~\ref{fig:char-owl-gpu} shows the power draw of all eight GPUs
during an Owl run at high concurrency. 
As each role is served by a separate serving instance and the roles differ widely in activity, this fine-grained
mapping translates role-level skew into uneven demand across GPUs. 
Four GPUs serving the busiest text and vision roles repeatedly
approach peak power, while the remaining GPUs spend most of the execution near idle $\sim$$20\%$. 
Under this static role-to-GPU mapping, lightly loaded devices retain substantial unused capacity.
Reclaiming this capacity requires collocating roles with
complementary activity on fewer GPUs. Differences in model size and
modality further constrain which roles can share a deployment or be
efficiently collocated, making the imbalance harder to eliminate. 

\begin{figure}[h]
\vspace{-2mm}
  \centering
  \includegraphics[width=\columnwidth]{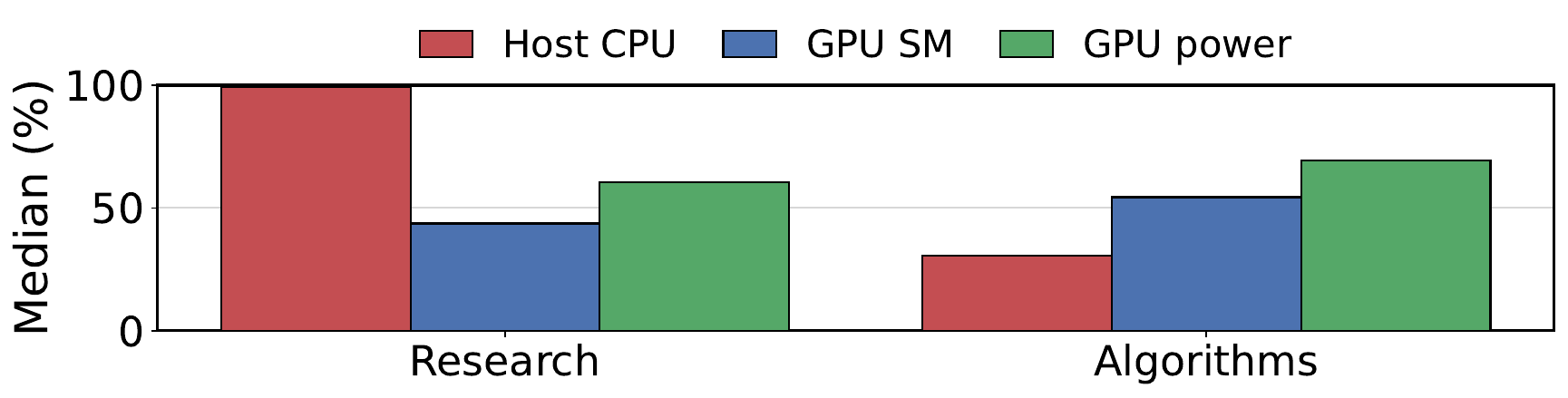}
  \vspace{-8mm}
  \caption{Resource utilization of CORAL on research vs. algorithmic task sets.}
  \label{fig:char-coral-tasks}
  \vspace{-2mm}
\end{figure}

\myparagraph{Workload diversity amplifies heterogeneity}
A workflow's taxonomy class alone does not determine its resource
demand.
Even within the same taxonomy class, resource demand varies with
the tasks an application solves and the tools it invokes.
Figure~\ref{fig:char-coral-tasks} holds the CORAL framework and workflow configuration fixed while
varying the task set. The same framework stresses opposite
resources depending on the task: research problems saturate the host CPU at
$99\%$ while leaving the GPU at $44\%$, whereas algorithmic problems exhibit the opposite balance, using only $31\%$ of the CPU but $55\%$ of the GPU. 

\begin{figure*}[t]
\vspace{-2mm}
  \centering
  \includegraphics[width=\textwidth]{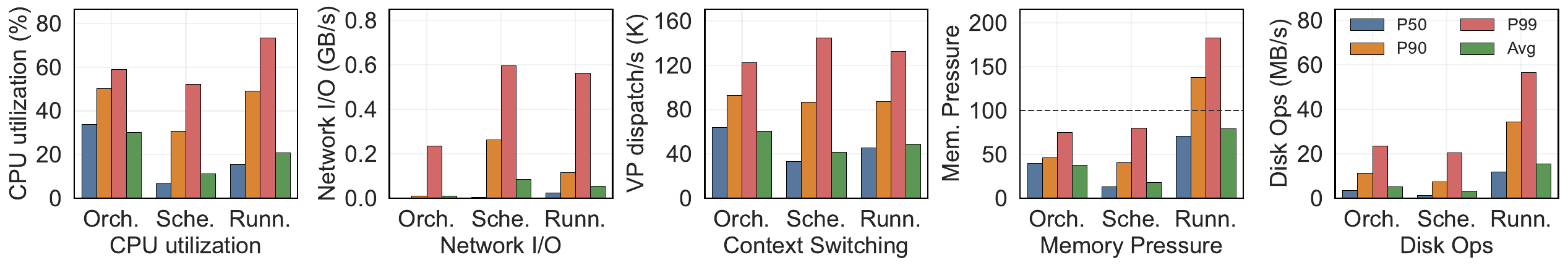}
  \vspace{-8mm}
  \caption{Per-role host metrics for schedulers, orchestrators, and runners in the
  fleet. The three roles have fundamentally different resource signatures.}
  \label{fig:cc-roles}
  \vspace{-1mm}
\end{figure*}

\begin{figure*}[t]
\vspace{-2mm}
  \centering
  \includegraphics[width=\textwidth]{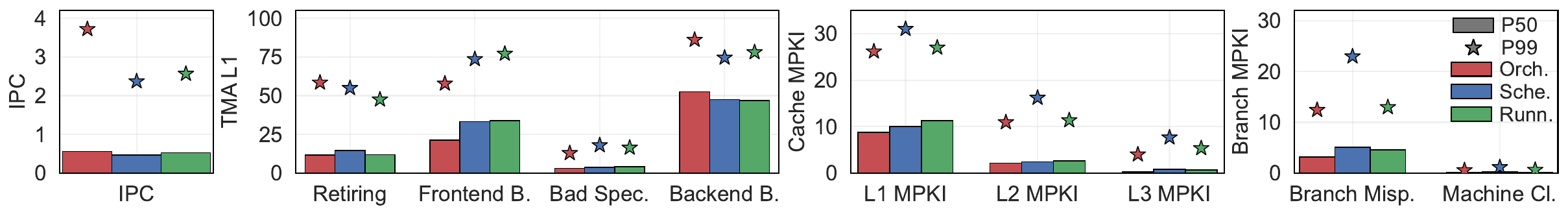}
  \vspace{-8mm}
  \caption{PMU measurements (top-down, cache, and branch) for the three host roles in the fleet. Low IPC, pipeline stalls, and high cache MPKI dominate.}
  \label{fig:cc-pmu}
  \vspace{-5mm}
\end{figure*}

\begin{figure}[h]
  \centering
  \includegraphics[width=\columnwidth]{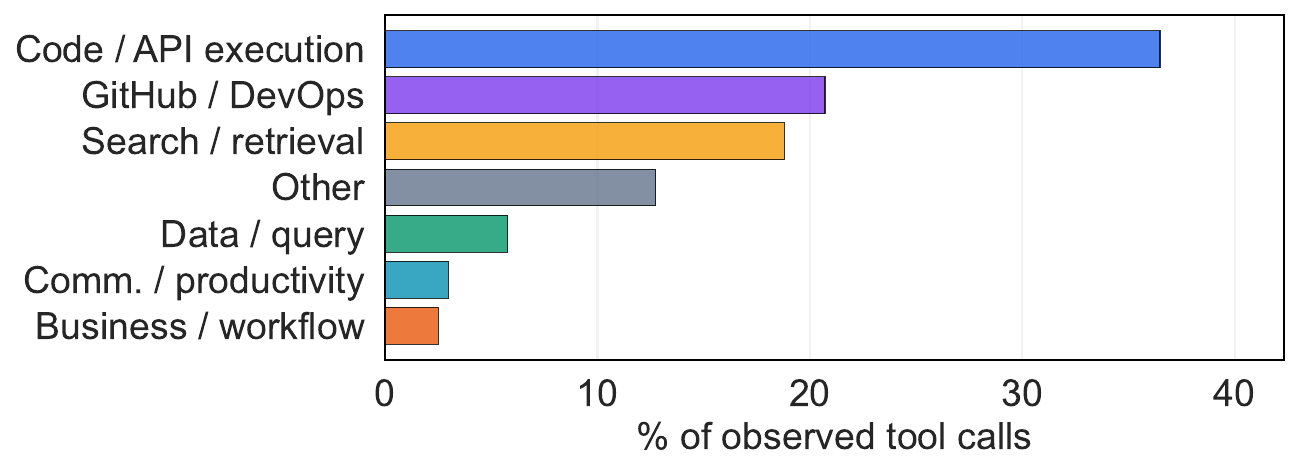}
  \vspace{-8mm}
  \caption{Fleet distribution of tool calls across functional categories.}
  \label{fig:cc-toolmix}
  \vspace{-2mm}
\end{figure}

\begin{figure}[h]
\vspace{-2mm}
  \centering
  \includegraphics[width=\columnwidth]{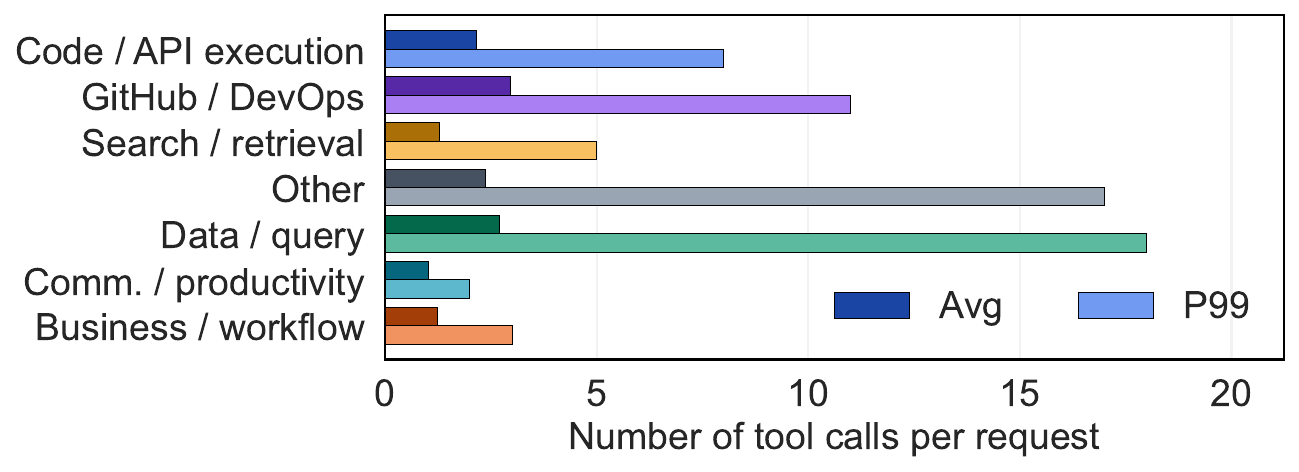}
  \vspace{-8mm}
  \caption{Per-request tool-call frequency by category (average and P99).}
  \label{fig:cc-callfreq}
  \vspace{-6mm}
\end{figure}

Across applications, Figure~\ref{fig:cc-toolmix} shows the distribution of tool
calls across functional categories in the fleet. The workload spans a
broad variety, from code execution, which is the single largest category, to DevOps, search and retrieval, databases, and
communication services, each exhibiting a distinct resource signature. 
Variation appears not only across categories but also across requests within a category. 
Figure~\ref{fig:cc-callfreq} shows
the number of concrete tool calls per active request in each category, reported as
the average and the 99th percentile. On average, a single request issues many
tool calls, and the tail is far heavier than the average and behaves distinctly across functional categories.
Consequently, diversity from tasks, applications, and tools further amplifies the heterogeneity and stresses the underlying hardware. 
No single CPU-to-GPU operating point characterizes the workload.

These results show the architectural challenges of running agentic workflows.
Host orchestration fragments execution, execution structure introduces bursty
load, model composition creates resource imbalance, and task and tool diversity
further broadens the operating space. 
We next examine how such heterogeneous work interacts with server architecture.

\subsection{Heterogeneous Work Stresses the Server Architecture}
\label{sec:uarch}
\label{sec:per-role}

The fragmentation and heterogeneity of agentic execution challenge
two assumptions underlying today's server architectures: resource
demand remains relatively stable over time, and host work can be
managed as a uniform pool. Conventional CPU-centric services
primarily stress the host, whereas monolithic LLM inference is largely
accelerator-bound. Agentic execution fits neither pattern. It alternates
between CPU and GPU phases, distributes work across roles
with distinct resource profiles, and multiplexes many short-lived tasks. 
Thus, static provisioning strands capacity,
while shared, role-agnostic scheduling introduces interference and
coordination overhead.

\begin{figure}[h]
\vspace{0mm}
  \centering
  \includegraphics[width=\columnwidth]{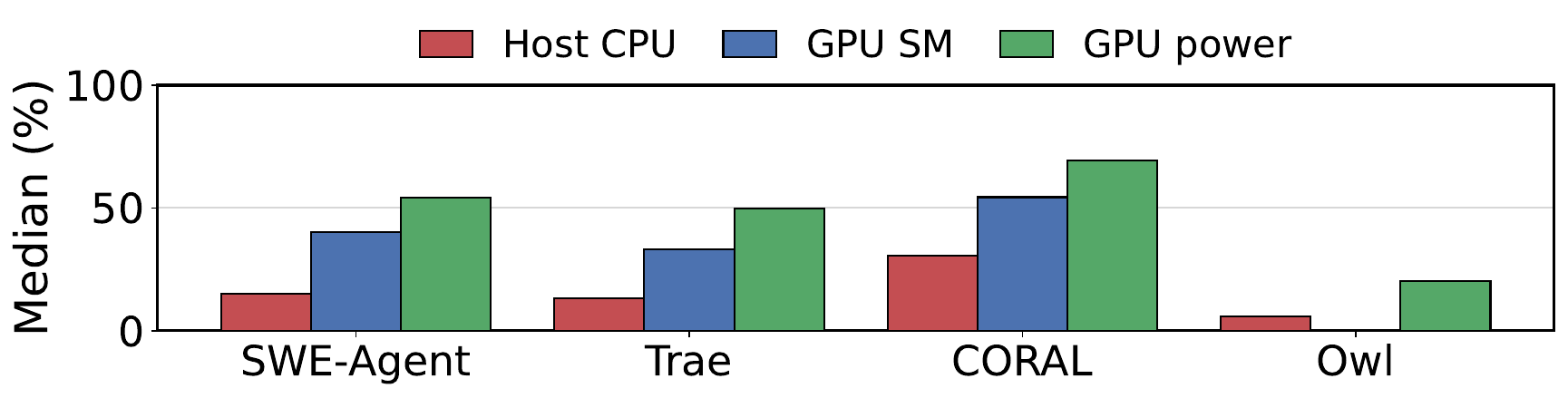}
  \vspace{-8mm}
  \caption{Median host CPU utilization, GPU SM-activity, and GPU power across
  frameworks on a single server.}
  \label{fig:char-util}
  \vspace{-6mm}
\end{figure}

\myparagraph{Fragmentation Strands CPU/GPU Resources}
Fragmentation causes CPU and GPU demand to alternate over time.
Figure~\ref{fig:char-util} shows the median host CPU utilization, 
GPU SM activity, and GPU power per framework. Both
processors remain underutilized on average: host CPU utilization ranges 
from $6\%$ (Owl) to $31\%$ (CORAL), while GPU SM activity stays below $55\%$. 
GPU power runs higher at up to $50$--$70\%$ of peak, 
but still leaves the accelerators far from saturated. 

The time-series show that low medians do not
reflect an absence of instantaneous demand. Instead, CPU demand falls
during accelerator-bound phases, while GPU activity drops during
host-side orchestration and tool execution. Provisioning for peak
wastes capacity between bursts, while provisioning for the
average leaves insufficient headroom when bursts arrive. Agentic
workloads strand capacity: neither the CPU nor GPU is
fully utilized in aggregate, yet each can become a transient bottleneck
on the workflow's critical path.

\myparagraph{Host roles break homogeneous provisioning}
The host side of an agentic system contains three distinct roles: schedulers that dispatch model requests,
orchestrators that manage workflow state and routing, and runners that execute
tools. Figure~\ref{fig:cc-roles} shows the resource signatures of production
hosts dedicated to these roles. The roles differ not only in their
overall utilization, but also in which resources they stress. 
For instance, a role that is CPU-heavy is light on the
network, and a role that dispatches at a high rate is light on disk. 
No single metric therefore captures their provisioning needs.

Mapping these roles onto a single homogeneous core pool leads to
inefficient provisioning, and simply adding more identical cores does
not resolve the underlying mismatch. Such a pool treats the roles as
interchangeable despite their distinct demand patterns and latency
requirements. Reserving enough cores for bursty runners wastes
capacity while tools are idle, whereas sharing those cores with the
control path allows tool bursts to delay inference dispatch and agent
handoffs. Agentic servers therefore require role-aware resource
allocation that isolates latency-critical coordination from bursty
execution and sizes each role's pool independently.

\myparagraph{Agent multiplexing impacts microarchitectural locality}
The mismatch extends below resource utilization into the microarchitecture itself. 
A host-wide PMU snapshot from production nodes provides fleet-scale evidence that these workloads can be microarchitecturally inefficient.
Figure~\ref{fig:cc-pmu} shows PMU measurements for cores serving the three host roles, including IPC, the
top-down breakdown of pipeline slots, cache misses per kilo-instruction, and branch
mispredictions. While IPC is low, the pipeline is dominated by
backend and frontend stalls rather than retiring instructions, and cache MPKI
is high, especially at the deeper levels of the hierarchy, with non-trivial branch
mispredictions on top. The host is thus starved on the memory subsystem and the
instruction front end, so its cores spend most of their cycles stalled rather than
doing useful work.

\begin{figure}[h]
\vspace{-3mm}
  \centering
  \includegraphics[width=\columnwidth]{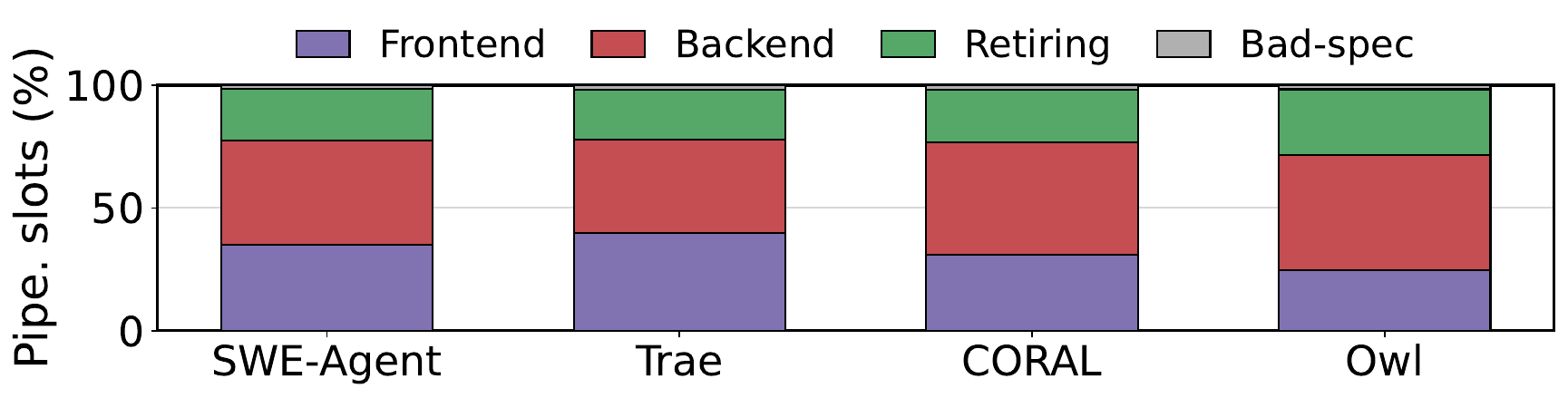}
  \vspace{-7mm}
  \caption{Top-down pipeline-slot breakdown across frameworks. The host is stalled
  on the memory subsystem or the instruction front end.}
  \label{fig:char-tma}
  \vspace{-3mm}
\end{figure}

Our controlled study provides further evidence consistent with
microarchitectural interference under agent multiplexing.
Figure~\ref{fig:char-tma} shows the top-down pipeline-slot
breakdown for the four frameworks. SWE-Agent, CORAL, and Owl are
primarily backend-bound, losing $43$--$47\%$ of pipeline slots to
backend stalls, while Trae is split more evenly between frontend and
backend stalls. IPC stays low, $1.2$--$1.6$. L1-data
MPKI reaches $14$--$20$, while L2 and branch-misprediction MPKI also
remain elevated. This behavior is consistent with interference in
stateful microarchitectural structures as many agent tasks time-share
the same cores, displacing one another's cache lines and
branch-predictor state. These results motivate bounding each task's
contention domain and preserving affinity, rather than freely
multiplexing all agent work across a shared core pool.

\begin{figure}[h]
\vspace{-4mm}
  \centering
  \includegraphics[width=\columnwidth]{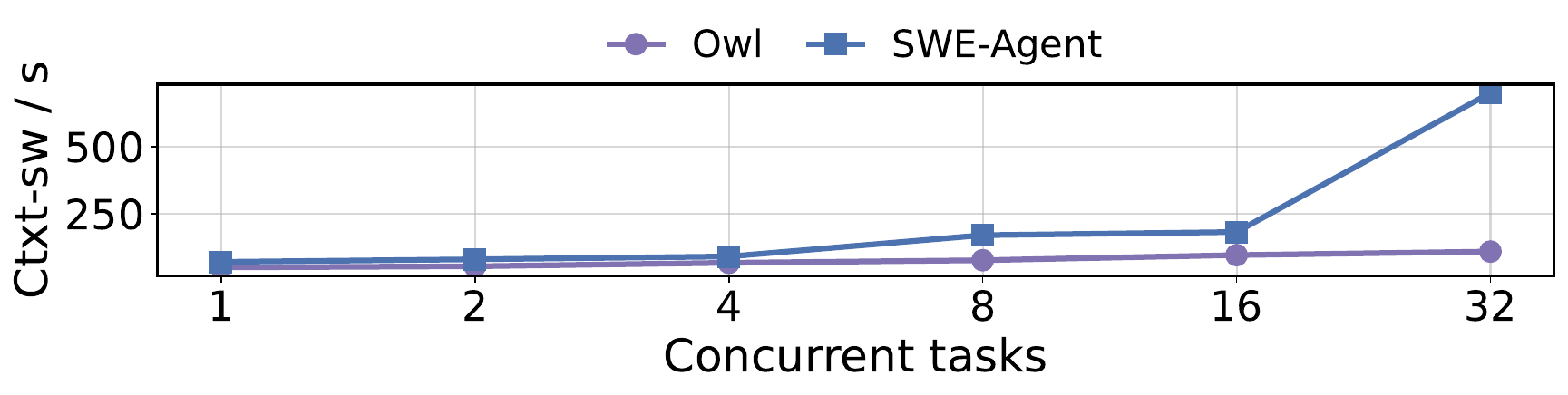}
  \vspace{-8mm}
  \caption{Involuntary context switches versus agent concurrency. Coordination
  cost grows steeply with the number of agents.}
  \label{fig:char-conc}
  \vspace{-2mm}
\end{figure}

\myparagraph{Coordination overhead limits scalability}
Finally, increasing agent concurrency creates additional host-side
coordination pressure. Unlike conventional inference, where higher
concurrency primarily improves accelerator utilization, adding
concurrent agents also increases host-side scheduling, dispatch, and
context-management activity. 
Figure~\ref{fig:char-conc} shows involuntary context switches, the signal of cores
being preempted under contention, as concurrency grows from 1 to 32 tasks.
\label{subsec:ctxsw}
For SWE-Agent, context switches grow from $71$ to $660$ per second, 
even though aggregate CPU utilization remains low.
The system therefore spends increasingly more effort coordinating agents 
rather than executing useful work. This behavior suggests that general-purpose 
operating system scheduling becomes insufficient at high agent concurrency 
and motivates dedicated hardware or runtime support for agent scheduling and context
management.

These results show that agentic AI stresses servers through a combination
of temporal fragmentation, resource heterogeneity, microarchitectural interference,
and growing coordination overhead. This is a poor match for static CPU/GPU partitioning and
undifferentiated core pools. Instead, agentic systems require
architectures that dynamically reclaim stranded resources, match hardware to
software roles, preserve locality under multiplexing, and reduce the cost of
coordination. 
\section{Server Design Principles and Case Studies}
\label{sec:design}
\label{sec:implications}

Our characterization yields concrete design principles for servers that run agentic AI.
We examine each through a case study in \emph{\system{}}, our prototype for commodity
servers. Agentic execution is fragmented, bursty, and heterogeneous,
whereas today's servers allocate resources statically and manage their cores as a
uniform pool.
\system{} closes this gap in two ways: it harvests the CPU and GPU capacity stranded by fragmentation,
and it pools and pins cores to match the differing demands of the scheduler, orchestrator, and runner.
\system{} tunes each principle to where the workload sits in the taxonomy, and to its current load
and task characteristics. 

\subsection{Harvesting Stranded CPU and GPU Cycles}
\label{subsec:harvest}

\subsubsection{Harvesting Idle CPU Cores}
\label{subsubsec:cpu-harvest}

The goal of CPU harvesting is to maximize per-server throughput while guarding the
latency of the agentic workflow. The characterization showed that runner cores
\textcolor{black}{exhibit substantial idle periods punctuated by sudden tool bursts}.
\system{} reclaims the idle cores by co-locating CPU-intensive traditional workloads, which we call
harvesters, on the same runner cores that the agent tools use. The difficulty is that the idle
capacity is not stable. A harvester that simply takes the idle cores steals them
during a tool burst and slows the agent workflow. \system{} thus lets harvesters share the
runner cores, harvesting aggressively while the runners are idle and retreating when
sustained runner demand appears.

\begin{figure}[h]
\vspace{-3mm}
  \centering
  \includegraphics[width=\columnwidth]{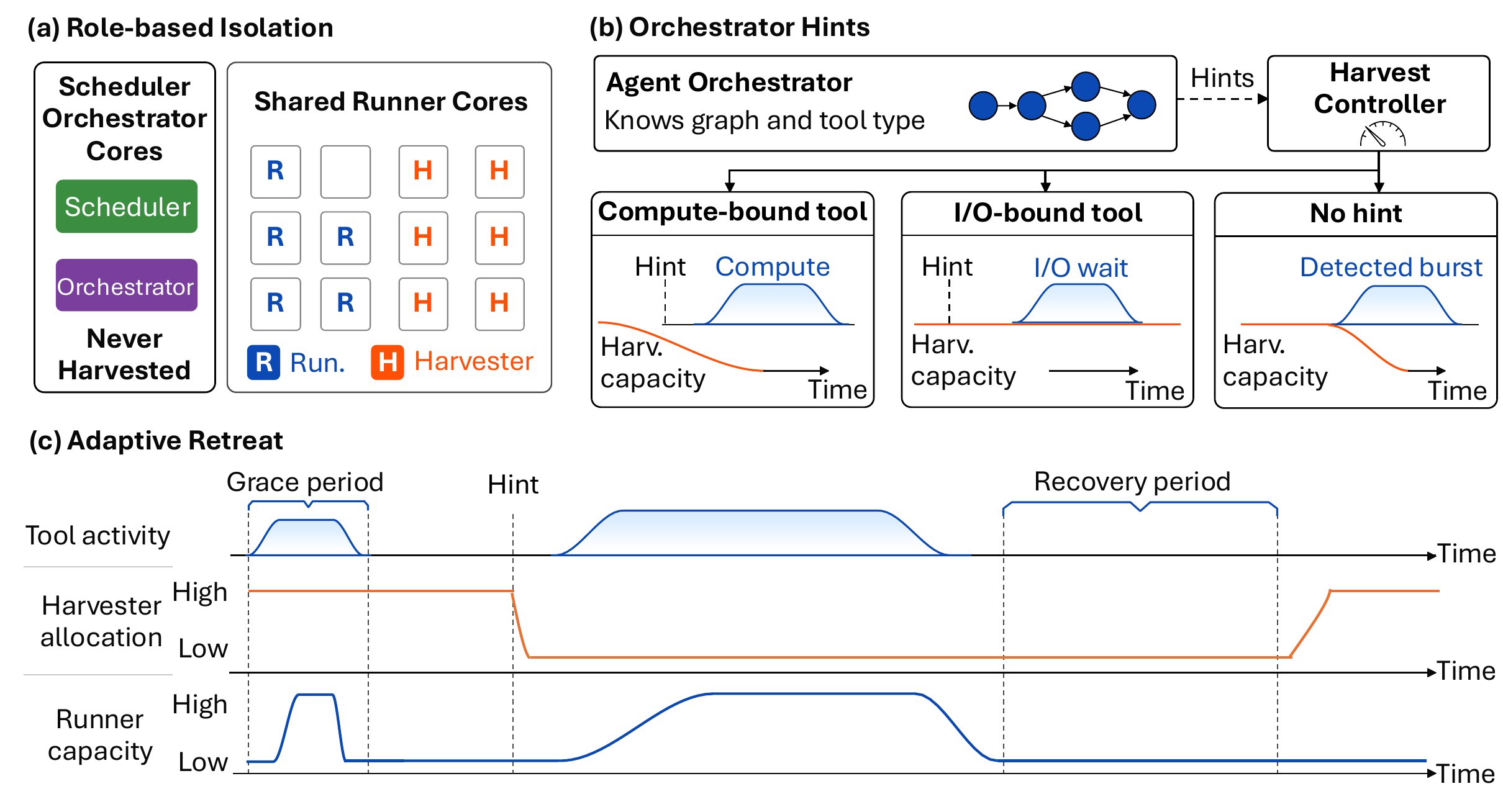}
  \vspace{-7mm}
  \caption{\system{} harvests idle runner cores in agentic workflows by collocating \emph{harvester} workloads and introducing three working mechanisms: a) role-based isolation, b) orchestrator hints, and c) adaptive retreat.}
  \label{fig:cpu-harv}
  \vspace{-2mm}
\end{figure}

Harvesting resources is well studied in datacenters, where prior systems
reclaim spare CPU cycles or memory from latency-critical services using historical utilization prediction and safe
throttling~\cite{historyHarvest,smartHarvest,memoryHVM,zhang2021faster,hardharvest}.
Agentic workloads \textcolor{black}{challenge} the assumptions behind these systems, because their idle
capacity \textcolor{black}{can disappear abruptly during overlapping tool bursts, so
history-based controllers may react too late}. \system{} differs in
three ways that are specific to agentic execution: it isolates cores by their role
so that harvesting never touches the latency-critical scheduler and orchestrator cores, it takes hints
from the orchestrator to anticipate bursts and to distinguish compute-bound from
I/O-bound tools, and it adapts its retreat policy to the workflow's position in the
taxonomy and to its load. Figure~\ref{fig:cpu-harv} illustrates the mechanism.

\vspace{2pt}\noindent\textbf{Role-based isolation.}
\system{} lets harvesters share the often-underutilized runner cores while
isolating the latency-critical control cores. These include scheduler cores
that dispatch inferences to the accelerators and orchestrator cores that route
work and state among agents; \system{} never places harvesters on either. This keeps the part
of the host that most directly affects agent latency free of interference.

\vspace{2pt}\noindent\textbf{Adaptive retreat.}
When \system{} detects a burst of tool executions on the runner cores, it first enters
a grace window. If the execution episode ends within the grace window, the harvester continues
unchanged,  \textcolor{black}{preserving harvester throughput when retreat would provide little benefit to the agent}.
If the burst persists, \system{} gradually throttles the harvester and eventually pauses it, limiting
contention while avoiding an abrupt loss of harvester throughput. After the episode ends,
\system{} also waits for a short recovery window before resuming the harvester,
merging closely spaced tool calls into one adjustment and preventing repeated throttling
and \textcolor{black}{resumption}.

\vspace{2pt}\noindent\textbf{Orchestrator hints.}
A reactive controller must observe a burst before it acts, \textcolor{black}{leaving a short window in which} 
tools and harvesters contend. \system{} closes this window when the
workflow is host-orchestrated: the orchestrator knows the workflow graph and warns
\system{} that a burst is about to start, so it takes action before
the tools run. The orchestrator also knows the tool types. When a tool is compute-bound, \system{}
\textcolor{black}{reduces the harvester's allocation}. When a tool is I/O-bound and 
\textcolor{black}{primarily waits} on an external service, \system{} keeps harvesting,
because the tool consumes little CPU.

\begin{table}[t]
\centering
\caption{Harvested throughput and agent slowdown under CPU harvesting at low and
high load. Values near $100\%$ mean the harvester approximately delivers its
standalone throughput.}
\vspace{-2mm}
\label{tab:harvest}
\footnotesize
\setlength{\tabcolsep}{5pt}
\renewcommand{\arraystretch}{1.15}
\begin{tabular}{lcccc}
\toprule
 & \multicolumn{2}{c}{Low load} & \multicolumn{2}{c}{High load} \\
\cmidrule(lr){2-3}\cmidrule(lr){4-5}
Framework & Harvest (\%) & Slow. (\%) & Harvest (\%) & Slow. (\%) \\
\midrule
SWE-Agent & 101.5 & 2.5 & 63.6 & 1.2 \\
Trae      & 100.1 & 1.4 & 73.3 & 4.3 \\
CORAL     &  94.7 & 1.0 &  1.0 & 9.9 \\
Owl       &  83.6 & 6.2 & 76.9 & 0.2 \\
\midrule
Average   &  95.0 & 2.8 & 53.7 & 3.9 \\
\bottomrule
\end{tabular}
\vspace{-6mm}
\end{table}

\vspace{2pt}\noindent\textbf{Reclaimed throughput.}
We evaluate CPU harvesting on the four frameworks, each co-located with a DCPerf~\cite{dcperf} harvester
\textcolor{black}{at low and high-loads}. The low-load setting runs a few concurrent agents and a light harvester, while the high-load setting runs many
concurrent agents and an intense harvester. We report two metrics: the agent
slowdown, which is the increase in the agent workflow's makespan relative to running
alone, and the harvested throughput, which is the fraction of the harvester's
standalone throughput that it delivers while co-located. Table~\ref{tab:harvest}
shows the results. At low load, the harvesters deliver $95\%$ of
their standalone throughput at under $3\%$ agent slowdown, and at high load,
they deliver $53.7\%$ on average at under $4\%$ slowdown. Averaged over the two
\textcolor{black}{operating} points, \textcolor{black}{the co-located harvesters deliver $74\%$ of their standalone
throughput, effectively recovering $0.74$ server-equivalents of additional
throughput}. \system{} \textcolor{black}{also} raises host CPU utilization by $31\%$, at a cost of
only $3.3\%$ higher agent latency.

\begin{figure}[h]
\vspace{-3mm}
  \centering
  \includegraphics[width=\columnwidth]{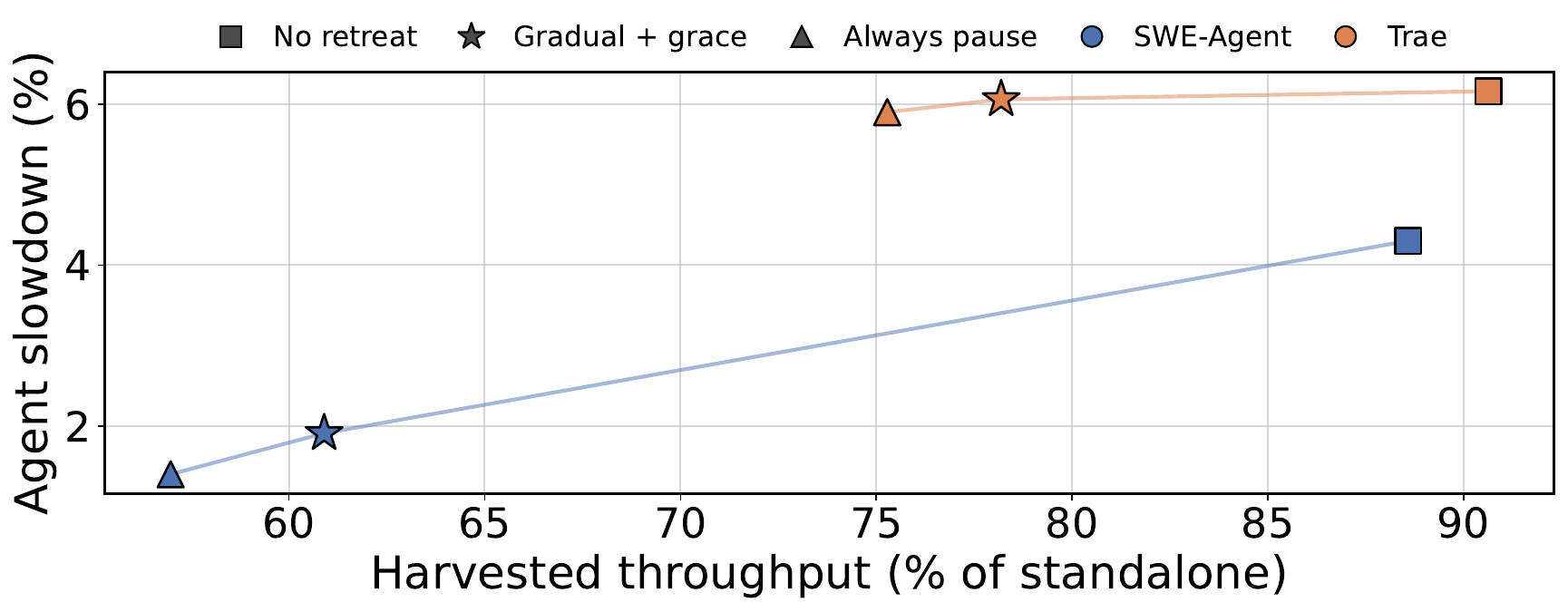}
  \vspace{-7mm}
  \caption{Harvested throughput versus agent slowdown at high load for three
  policies, with two different agentic frameworks. 
  }
  \label{fig:harvest-policy}
  \vspace{-2mm}
\end{figure}

\vspace{2pt}\noindent\textbf{Policy trade-off.}
Figure~\ref{fig:harvest-policy} shows, for two frameworks at high load, the
harvested throughput and the agent slowdown of three policies: never retreating,
\textcolor{black}{pausing the harvester whenever a tool runs}, and the gradual retreat with a grace
period.
Never retreating delivers the most throughput but the
largest slowdown. Always pausing gives the least slowdown but the least
throughput. \textcolor{black}{The gradual policy avoids most of the slowdown caused
by no retreat while delivering more throughput than always pausing.} \system{}
thus uses the gradual, grace-guarded retreat at high load and the no-retreat
policy at low load. Two cases bound the technique. A parallel, tool-saturated
workflow keeps its cores busy almost continuously, so little is harvestable and the
delivered throughput is low, as CORAL shows at high load. A workflow like Owl whose tools
mostly wait on the network is over-protected by tool-triggered retreat, which
is the case that the orchestrator's tool-type hint is designed to fix. 
Without type information, an all-tool retreat policy retains only 37.7\% standalone throughput; 
with tool-type hints, it retains 80\% at 0.2\% average agent slowdown.

\vspace{2pt}\noindent\textbf{Adapting to the taxonomy and load.}
How much and how early \system{} retreats depends on the workload's
taxonomy and on the load. Along the execution-structure axis, a sequential workflow
issues isolated, well-separated tool bursts, so most bursts end within the grace period
and the harvester keeps most of its throughput. A parallel workflow
\textcolor{black}{can issue concurrent bursts across multiple agents}, so the bursts overlap and leave little idle time. 
In the extreme, a parallel and tool-saturated
workflow (e.g., CORAL at high load) leaves almost no idle capacity, so \system{} 
disables harvesting entirely.  
Along the orchestration axis, a
host-orchestrated workflow supplies the proactive and typed hints above, so
\system{} retreats selectively for compute-bound tools, whereas an
LLM-orchestrated workflow gives no hints and \system{} falls back to reactive
detection with a conservative grace period. The load sets the operating point: at
low load, the interference is small and \system{} harvests aggressively, and at high
load, 
\system{} applies the gradual, grace-guarded retreat to bound the agent's
latency.

\subsubsection{Harvesting Idle GPU \textcolor{black}{Capacity}}
\label{subsubsec:gpu-harvest}

\begin{figure}[h]
\vspace{-3mm}
  \centering
  \includegraphics[width=\columnwidth]{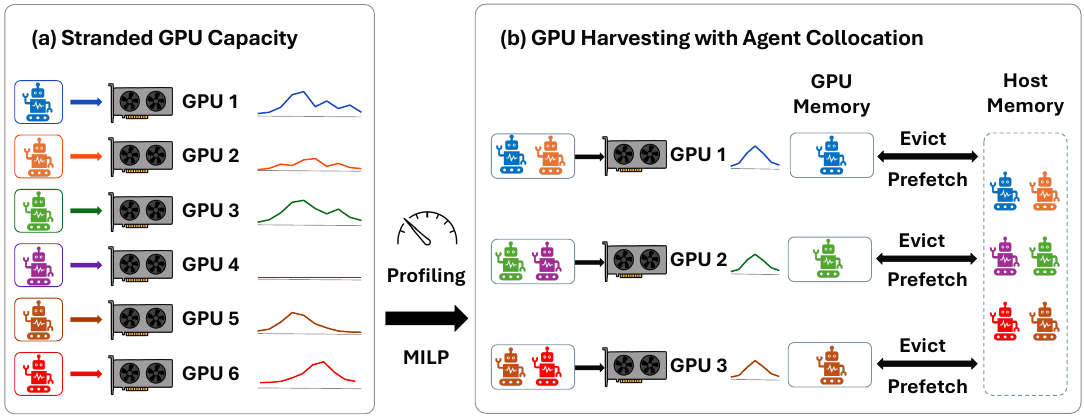}
  \vspace{-7mm}
  \caption{GPU layout: a) a static role-dedicated deployment that strands GPU
  compute, and b) \system{}'s GPU harvesting, which consolidates agents onto shared
  serving instances and frees GPUs.}
  \label{fig:gpu-harv}
  \vspace{-3mm}
\end{figure}

The goal is to serve the same agents on fewer GPUs, so that each
accelerator does useful work instead of idling. The characterization showed that the
accelerator pool is underused and imbalanced under a \textcolor{black}{static, role-dedicated deployment}.
Some devices approach saturation while others sit near idle, because \textcolor{black}{roles} differ in
load and are often not active at the same time.
\system{} reclaims this stranded
capacity in two steps. It first \emph{consolidates} agents by \textcolor{black}{co-locating several on a
shared serving instance, so that one instance serves many agents, raising each GPU's utilization at a fixed GPU count}. It then \emph{harvests} by
\textcolor{black}{reducing the GPU count itself}, returning the freed GPUs to a shared pool for other work. 
The challenge is that the agents' combined \textcolor{black}{memory footprint may
exceed device capacity, while roles that are active simultaneously can
contend for GPU compute.} Figure~\ref{fig:gpu-harv} illustrates the mechanism.

\vspace{2pt}\noindent\textbf{Oversubscribing GPU memory.}
\system{} oversubscribes a GPU by \textcolor{black}{assigning it more agent state than
can reside simultaneously}. It
keeps the active agents resident and stages the rest in host memory. Swapping this
state in on demand would stall the accelerator, so \system{} hides the cost when it
can. When the workflow is host-orchestrated, the orchestrator knows which agent runs
next, and \system{} prefetches that agent's state from host memory before it is needed,
overlapping the transfer with the current agent's work.

\vspace{2pt}\noindent\textbf{Taxonomy-guided collocation.}
Which agents to place together \textcolor{black}{depends on} the taxonomy and how much state they share.
Agents that share large amounts of state, \textcolor{black}{including model weights and cached prompt prefixes},
are \textcolor{black}{efficient} to collocate, because the shared portion stays resident only once. A homogeneous
workflow is the ideal case, where every agent \textcolor{black}{uses the same base model} and only the
smaller per-agent state must be \textcolor{black}{managed separately}.
Agents that are never active at the same time can share a device without contending for compute,
which a sequential workflow guarantees since it activates one agent at a time. \system{} also pairs a heavily used
agent with a lightly used one rather than stacking two hot agents on the same device, so
that collocated agents fill each other's idle periods instead of competing.

\vspace{2pt}\noindent\textbf{Solving the allocation.}
The best layout depends on the models, the workflow, and the load, all of which shift
over a run, so \system{} recomputes it dynamically. 
It casts
the placement as a mixed-integer linear program over profiling data. The program
decides, for each agent, how many GPUs it receives and its tensor-parallel
configuration, and which agents to collocate, subject to the memory and contention
constraints above. The result is a GPU assignment that tracks the workload and keeps
the accelerators busy without slowing the agents.

\begin{table}[t]
\centering
\caption{GPU consolidation (con.) and harvesting (har.) under high load, relative to a \textcolor{black}{role-dedicated deployment} baseline.} 
\vspace{-3mm}
\label{tab:gpu-harvest}
\footnotesize
\setlength{\tabcolsep}{3pt}
\renewcommand{\arraystretch}{1.15}
\resizebox{\columnwidth}{!}{%
\begin{tabular}{lcccccc}
\toprule
  Framework & GPUs red. & T-put. & Tasks/h
            & Tail lat. & Energy eff. & KV hit \\
  \midrule
  Owl (con.)   & --     & $+106\%$ & $+36\%$
               & $3.8\times$ & $+42\%$ & $+68\%$ \\
  Owl (har.)   & $33\%$ & $+82\%$  & $+22\%$
               & $2.5\times$ & $+51\%$ & $+75\%$ \\
  CORAL (con.) & --     & $+15\%$  & $+28\%$
               & $5.3\times$ & $+20\%$ & $+8\%$ \\
  CORAL (har.) & $50\%$ & $-71\%$  & $-55\%$
               & $16\times$~worse & $-53\%$ & $-17\%$ \\ 
\bottomrule
\end{tabular}%
}
\vspace{-5mm}
\end{table}

\vspace{2pt}\noindent\textbf{Reclaimed capacity.}
Table~\ref{tab:gpu-harvest} reports GPU harvesting on two workflows, Owl and CORAL,
relative to serving each agent on its own GPUs. For Owl, \system{} frees a third of the
GPUs while \emph{raising} generation throughput by $82\%$, completing $22\%$ more tasks
per hour, cutting tail latency by $2.5\times$, and improving energy efficiency by
$51\%$. The benefit holds across the load range, so the harvested layout is safe as a
static choice. At a fixed GPU budget, the same \textcolor{black}{consolidation mechanism instead improves} performance,
raising throughput by $106\%$ and cutting tail latency by $3.8\times$. Consolidation
improves throughput even as it frees GPUs because collocated agents that run the same
model share a single copy of its weights. This removes the duplicate weight copies that
a \textcolor{black}{role-dedicated deployment} keeps, \textcolor{black}{freeing memory for a larger KV cache. Together with
cross-agent prefix reuse, this additional capacity increases the measured
KV-cache hit rate and serving capacity.}
\textcolor{black}{Higher tensor parallelism can improve throughput by
amortizing weight reads, but introduces a synchronization--communication
trade-off, e.g., in CORAL, TP8 does not outperform the role-dedicated baseline,
while two TP4 instances do.}

\vspace{2pt}\noindent\textbf{Adapting to the taxonomy and load.}
How much \system{} can harvest depends on where the workload sits in the taxonomy, and
the execution structure sets the headroom. When agents run at different times rather than
in parallel, their busy periods do not overlap, so many GPUs sit idle at any instant.
\system{} can then time-share those agents onto fewer devices and reclaim whole GPUs.
Owl behaves this way, as its agents activate unevenly and leave much of the pool
stranded, so harvesting frees a third of the GPUs. A parallel workflow such as CORAL
instead \textcolor{black}{keeps its agents active concurrently, leaving little
capacity stranded}, and harvesting GPUs there \textcolor{black}{costs throughput that can
exceed the fraction of GPUs removed}. \textcolor{black}{Such a workflow should
therefore not be harvested.} It still
consolidates the agents at a fixed GPU budget, where sharing weights and cache raises
throughput by $15\%$ and cuts tail latency by $5.3\times$ at high load.  Load sets the
operating point: the gains are largest at high concurrency, where stranding and
contention are worst, and shrink at low load. \system{} thus harvests GPUs where the
\textcolor{black}{taxonomy and load} leave them stranded, and otherwise consolidates only to raise throughput.

\subsection{Role-Aware CPU Core Pooling and Pinning}
\label{subsec:pooling}

In a conventional role-agnostic deployment, the scheduler, orchestrator, and runner share one CPU pool, despite performing
different kinds of work. The scheduler and the orchestrator are latency-critical:
they gate every inference and handoff, so response time matters. The runner
instead has highly variable demand: its tools sit idle, then burst. These demands pull in opposite
directions, so the roles cannot be scaled as one bundle. A shared pool sized for
bursts wastes cores when the runner is idle. A pool sized for the common case has no
room for a burst. The tools then steal cores from the control roles the agent is
waiting on. Sharing a pool also reduces the effectiveness of microarchitectural structures:
many tasks and tools
crowd onto the same cores and switch in and out constantly, polluting the caches and
branch predictors.

\begin{figure}[h]
\vspace{-2mm}
  \centering
  \includegraphics[width=\columnwidth]{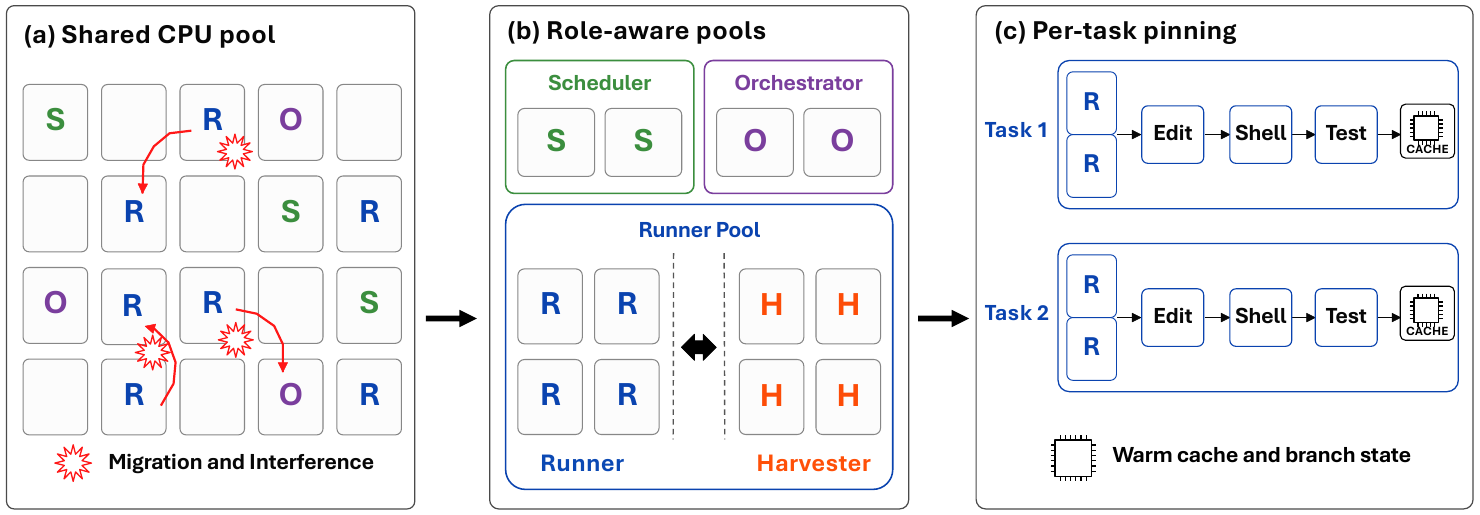}
  \vspace{-7mm}
  \caption{CPU core layout: a) a shared pool where all roles contend for cores,
  causing interference and migration, b) role-aware pools, and c) per-task pinning
  within the runner pool.}
  \label{fig:pooling}
  \vspace{-2mm}
\end{figure}

\system{} responds by giving each role its own pool. It separates the control roles
from the runner, so the two no longer compete and each is sized on its own. Within the
runner, it pins each task to a narrow set of cores, limiting tool migration and cache thrashing. The challenge is that the layout cannot be fixed once, unlike
with the traditional datacenter services~\cite{mosaic}. The runner
swings between idle and saturation, so the partition that fits one workload is wrong
for the next. To solve this, \system{} builds on its harvesting mechanism 
and extends its role-awareness inward to the
agent's cores, dynamically resizing the pools. 
Figure~\ref{fig:pooling} shows the mechanism. 

\vspace{2pt}\noindent\textbf{Right-sizing the role pools.}
\system{} places the scheduler and the orchestrator on small, dedicated pools of a
few cores each. Both roles are latency-critical but with low and stable demand, so isolation keeps inference dispatches and handoffs off the contended runner cores. Thus, a tool burst cannot directly contend for their cores. Profiling the scheduler
plane shows that its footprint is small and stable and grows with the serving
configuration rather than with the agent load, so \system{} sizes each control pool
just above its measured demand.

The remaining cores form the bounded runner pool for bursty, mostly idle tools. 
Compaction therefore reduces scheduling overhead because the runner role incurs the most context switches. On a wide set of cores, the OS repeatedly migrates the many
short-lived tool processes, and each migration reschedules the process and reloads its
state on a new core, whereas a bounded pool holds them on a limited set of cores.
Compaction also keeps the runner's bursts from spilling into the control pools, and
the cores left outside it become clean, contiguous capacity that harvesting can
reclaim.

\vspace{2pt}\noindent\textbf{Pinning to restore locality.}
Sizing the runner pool is not enough, as many agents still share the same cores. \system{} therefore
pins each task, or a small group of tasks, to a narrow set of cores within the
runner pool. A narrow contention domain restores the cache and branch locality that
the agents would otherwise destroy, and it bounds how far one task's burst can
interfere with its neighbors.

\system{} pins by task rather than by tool type. A task is one agent's assignment,
such as resolving a single issue. In our coding workloads, over its lifetime it runs a whole sequence of
tools against the same files and repository state (e.g., editing a file,
running a shell command, and executing a test). Pinning by task keeps that entire
sequence on a set of cores. Pinning by tool type would instead gather every
edit across all tasks onto one set of cores and every test onto another. The mix of
active tools shifts continuously over a run, so a static per-tool partition cannot
track demand and leaves some tool classes starved while others sit idle. A per-task
partition instead follows the work, as each task carries its own tools with it.
It also preserves locality, as the tools within a task operate on the same
working set. 

\vspace{2pt}\noindent\textbf{Adapting to the taxonomy and load.}
The pool layout follows where the workload sits in the taxonomy. Along the model-composition dimension, a homogeneous workflow can share one scheduler pool, whereas a heterogeneous workflow requires one per serving backend. Along the execution-structure dimension, a parallel, high-concurrency workflow gains the most from
fine-grained per-task pinning, because many tool bursts overlap and contend, while a
sequential workflow contends little and needs coarse pinning at most. The load sets
the pool sizes, so \system{} widens the runner pool as concurrency rises and returns
cores to harvesting when the runner is quiet.

\vspace{2pt}\noindent\textbf{Evaluation setup.} 
We increase load by the number of concurrent agent tasks,
$C \in \{1, 8, 32\}$, on a fixed core budget: $C{=}1$ and $C{=}8$ use one $24$-core
NUMA node, and $C{=}32$ uses $48$ cores across two nodes. Compaction confines tools to a bounded runner pool rather than the full core budget. We compare \emph{Shared}, which places the tool and serving planes in one CPU set; \emph{Split}, which assigns them separate pools on the same node; and \emph{Isolate}, which moves the serving plane to another NUMA node while preserving the total core budget. Within the runner pool we sweep pinning from one wide shared pool to a few tasks per pool, one shared neighbor, and private per-task cores. 

We report reclaimed CPU as CPU-seconds relative to the unpartitioned baseline. For performance, we report batch completion time—the wall-clock time of the whole run—and per-lane metrics. We break performance down per \emph{lane}, which is one task's execution timeline:
its full-lane span is the task's end-to-end time, and its tool sum is the cumulative execution time of its tool calls. We report per-lane metrics as the median and maximum across lanes. To size
the scheduler pool, we separately run a pure-serving microbenchmark that shrinks the
serving core budget while measuring output throughput.

\vspace{2pt}\noindent\textbf{Right-sizing the scheduler pool.}
The serving plane's host demand is small and stable. It uses $10.3$ cores in
total across its control processes, and this holds across frameworks.
Throughput then saturates
sharply with the core budget: a pool of $11$ cores already retains $98.5\%$ of
standalone throughput, while $16$ or more cores add almost nothing.
\system{} therefore sizes the scheduler pool just above this measured demand. Doing so
retains $99\%$ of serving throughput and frees the rest of the machine for the
runner and for harvesting.

\vspace{2pt}\noindent\textbf{Compacting the runner pool.}
Table~\ref{tab:compaction} reports runner compaction at two
concurrency levels. Compaction reclaims a large share of the tool
footprint while leaving the serving plane untouched. It cuts tool
CPU-seconds by $25$ to $46\%$ and changes serving CPU by under $2\%$, lowering runner usage from $15.6$ to $11.5$ cores at high concurrency. Separating the
roles matters: the split and isolated layouts save 4.1--8.5 more points of tool CPU
than a shared pool, and they avoid the cross-role contention that slows the shared
pool at $C{=}8$. The reclaimed capacity comes at no end-to-end cost, since batch
completion changes by under $1\%$. Overall,
compaction turns scattered tool cores into clean
harvestable capacity. 

\begin{table}[t]
  \centering
  \caption{Runner-pool compaction relative to the unpartitioned
  baseline (\%). Positive means less CPU work or faster execution (median/max across lanes).}
  \vspace{-3mm}
  \label{tab:compaction}
  \footnotesize
  \setlength{\tabcolsep}{4pt}
  \renewcommand{\arraystretch}{1.15}
  \begin{tabular}{clrrrr}
    \toprule
    $C$ & Placement & Tool CPU & Serve CPU & Tool time & Batch \\
    \midrule
     8 & Shared  & 40.1 & -0.2 & -2.4 / -1.2 & -0.3 \\
     8 & Split   & 44.2 &  1.4 &  2.3 /  4.2 &  0.2 \\
     8 & Isolate & \textbf{46.0} &  1.0 &  1.7 /  5.6 &  \textbf{1.0} \\
    \midrule
    32 & Shared  & 25.2 &  0.9 &  3.6 / 17.0 &  0.6 \\
    32 & Split   & 29.7 &  1.8 &  \textbf{7.0} / 20.1 &  0.2 \\
    32 & Isolate & \textbf{33.7} &  0.5 &  4.4 / \textbf{21.1} &  0.3 \\
    \bottomrule
  \end{tabular}
  \vspace{-5mm}
\end{table}

\vspace{2pt}\noindent\textbf{Locality from pinning.}
Table~\ref{tab:pooling} reports the effect of narrowing each task's contention
domain at high concurrency, relative to a wide shared runner pool. The benefit shows
up most on the slowest lanes. The best layout, two tasks per pool,
shortens tool time by up to $12.8\%$ on the tail, cuts tool CPU-seconds by
$34\%$, and speeds up batch completion by $4.7\%$. Partitioning helps because a narrow
contention domain bounds how far one task's burst spills into its neighbors, so the
tasks that would otherwise suffer the worst interference recover the most. Narrowing
too far backfires, 
e.g., fully private cores do worse than the
shared pool on batch time, because they remove the slack a task needs to absorb its
own bursts. Partitioning by tool class instead of by task gives no benefit,
which confirms that the layout must follow the shifting per-task tool mix. 

\begin{table}[h]
\vspace{-3mm}
  \centering
  \caption{Effect of narrowing each task's contention domain at high concurrency,
relative to a wide shared runner pool (\%). Positive values indicate
faster execution. Full-lane and
tool-sum are median/maximum across lanes.}
  \vspace{-3mm}
  \label{tab:pooling}
  \footnotesize
  \setlength{\tabcolsep}{5pt}
  \renewcommand{\arraystretch}{1.15}
  \begin{tabular}{lrrr}
    \toprule
    Contention domain & Batch & Full lane & Tool sum \\
    \midrule
    16 tasks / pool      &  2.58 & 1.52 / 11.49 & 1.69 / 11.47 \\
    4 tasks / pool       &  2.39 & 2.38 / 11.19 & 2.63 / 12.15 \\
    2 tasks / pool       &  \textbf{4.73} & \textbf{3.30} / 11.58 & \textbf{3.24} / 12.81 \\
    1 neighbor (ring)    & -0.65 & 3.29 / 12.76 & 2.43 / 13.11 \\
    Private cores        & -1.01 & 1.53 / 11.47 & 2.79 / 12.26 \\
    \bottomrule
  \end{tabular}
  \vspace{-2mm}
\end{table}

\vspace{2pt}\noindent\textbf{Adapting to the workloads.}
Tool workloads exhibit diverse resource profiles and differ both in
how aggressively they can be compacted and in how much they benefit from
compaction. Owl on GAIA~\cite{gaia}  spends most of its
tool time waiting on network I/O and has low baseline CPU demand. In a separate Owl experiment at
$C{=}32$, reducing its total core budget from 96 to a 24-core isolated layout
preserves performance  
and reduces
average runner usage by 4.8\%. 
Although this reduction
is smaller than for SWE-Agent, the tighter layout cuts peak runner usage from
70.1 to 14.1 cores, reducing interference with co-located workloads.

Harvesting and role-aware pooling complete \system{}'s CPU management.
Harvesting fills idle cores with useful work, while pooling lays out the
agent's own roles so that the control path stays fast, the tools keep their
locality, and the reclaimed capacity forms clean pools for the harvester to use.

\subsection{Implications for Future Hardware}
\label{subsec:future-hw}

Our characterization also points to where
future silicon could help agentic AI most. We highlight three directions.

\vspace{2pt}\noindent\textbf{Hardware support for scheduling and context switching.}
Coordination is a first-class cost: context switches grow with concurrency despite low utilization, and the runner incurs the
most context switches among the three roles. \system{}'s pooling and pinning bound but cannot
eliminate this cost, since the OS scheduler still dispatches every short-lived tool and handoff. A
dedicated engine that offloads agent scheduling and context switching from the
OS~\cite{micromanycore,altocumulus}, holding agents and their state
in hardware queues, would turn concurrency into throughput.

\vspace{2pt}\noindent\textbf{Hardware partitioning of microarchitectural structures.}
The host's low IPC comes from contention for stateful structures: many similar agents on
shared cores can evict one another's cache lines and branch history, raising MPKI.
\system{} recovers some locality by pinning tasks to narrow core sets, but this is
coarse, dedicating whole cores and wasting capacity when a task underuses them.
Partitioning the shared structures directly~\cite{mosaic,microsliced}, such as way-partitioned or agent-tagged
caches, TLBs, and branch predictors, would isolate agents \emph{within} a core more finely
than software affinity.

\vspace{2pt}\noindent\textbf{Heterogeneous and morphable cores.}
The three roles have different needs: the scheduler and orchestrator are
latency-critical and coordination-bound, and the runner is throughput-oriented and
compute-heavy, so a uniform core serves all three poorly. Heterogeneous per-server cores~\cite{biglittle,singleisaheterogeneity,phaseweave}
would fit each better, pairing energy-efficient cores for orchestration with
high-performance cores for tool execution, with the scheduling engine above.
As the role mix dynamically shifts, static heterogeneity is a
compromise; morphable cores~\cite{morphcore,duplexity} that reconfigure their resources could track these shifts, 
extending \system{}'s software adaptation to hardware.

\section{Conclusion}
\label{sec:conclusion}

Agentic AI is becoming a major datacenter workload, yet its architectural
implications have remained largely unexamined. This paper took a first step by 
systematically 
characterizing the agentic workflows at production scale. We found that agentic execution  
fragments tasks into short, interdependent,
host-orchestrated steps. They leave CPU and GPU underutilized on
average yet bursty and imbalanced at any instant, while placing heavy pressure on
the host microarchitecture.
Guided by
these findings, we designed \system{}, an agent-native server runtime that
reclaims the stranded compute on both the host and the accelerator, provisions and
schedules cores by role,
and adapts to the diversity of agentic workloads.


\bibliographystyle{IEEEtranS}
\bibliography{refs}

\end{document}